\pdfoutput=1

\documentclass[11pt]{article}

\usepackage[preprint]{acl}

\usepackage{times}
\usepackage{latexsym}

\usepackage[T1]{fontenc}

\usepackage[utf8]{inputenc}

\usepackage{microtype}

\usepackage{inconsolata}

\usepackage{graphicx}

\usepackage{hyperref}
\usepackage{url}
\usepackage{graphicx}
\usepackage{xcolor}
\usepackage{booktabs}
\usepackage{amsmath}

\usepackage{adjustbox}
\usepackage{subcaption}
\usepackage{amssymb}
\usepackage{pifont}

\usepackage{listings}
\newcommand{\xmark}{\ding{55}}%
\newcommand{\improve}[1]{($\textcolor{green}{\uparrow #1}$)}
\newcommand{\worse}[1]{($\textcolor{red}{\downarrow #1}$)}

\usepackage[utf8]{inputenc}

\usepackage{microtype}

\usepackage{inconsolata}

\title{FoREST: \textbf{F}rame \textbf{o}f \textbf{R}eference \textbf{E}valuation in \textbf{S}patial Reasoning \textbf{T}asks}

\author{Tanawan Premsri \and Parisa Kordjamshidi\\
          Department of Computer Science  and Engineering \\
       Michigan State University\\
       \{premsrit, kordjams\}@msu.edu}

\begin{document}
\maketitle
\begin{abstract}
Spatial reasoning is a fundamental aspect of human intelligence. 
One key concept in spatial cognition is the Frame of Reference (FoR), which identifies the perspective of spatial expressions. 
Despite its significance, FoR has received limited attention in AI models that need spatial intelligence. There is a lack of dedicated benchmarks and in-depth evaluation of large language models (LLMs) in this area.
To address this issue, we introduce the \textbf{F}rame \textbf{o}f \textbf{R}eference \textbf{E}valuation in \textbf{S}patial Reasoning \textbf{T}asks (FoREST) benchmark, designed to assess FoR comprehension in LLMs.
We evaluate LLMs on answering questions that require FoR comprehension and layout generation in text-to-image models using FoREST.
Our results reveal a notable performance gap across different FoR classes in various LLMs, affecting their ability to generate accurate layouts for text-to-image generation. 
This highlights critical shortcomings in FoR comprehension.
To improve FoR understanding, we propose Spatial-Guided prompting, which improves LLMs’ ability to extract primitive spatial concepts and relations. 
Our proposed method improves overall performance across spatial reasoning tasks.
\end{abstract}

\section{Introduction}
\begin{figure}[t]
    \centering
    \includegraphics[width=0.75\columnwidth]{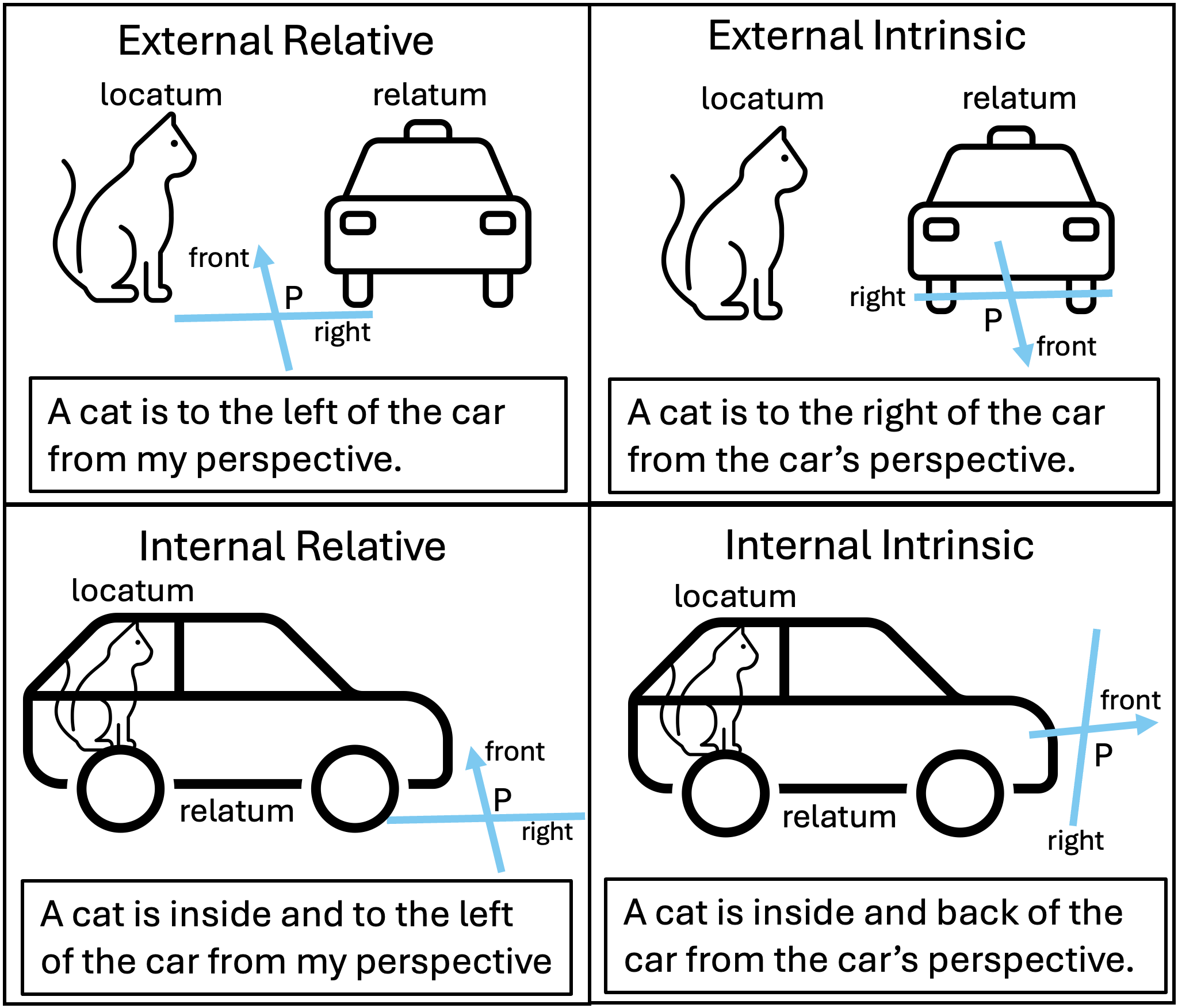}
    \caption{Illustration of FoR classes. The cat is the locatum, the car is the relatum, and arrows denote the perspective.}
  \label{fig:FoR_classes}
\end{figure}

Spatial reasoning plays a significant role in human cognition and daily activities. 
It is also a crucial aspect in many AI problems~\cite{, KordjamshidiMoensPustejovsky2026}, including language grounding~\citep{zhang2022lovis, 10610443}, navigation~\citep{zhang-kordjamshidi-2023-vln, yamada2024evaluatingspatialunderstandinglarge}, computer vision~\citep{liu2023visualspatialreasoning, chen2024spatialvlmendowingvisionlanguagemodels}, medical domain~\citep{gong20233dsamadapter}, and image generation~\citep{cho2023visualprogrammingtexttoimagegeneration, gokhale2023benchmarkingspatialrelationshipstexttoimage}.
One key concept in spatial reasoning is the Frame of Reference (FoR), which identifies the perspective of spatial expressions. FoR has been studied extensively in cognitive linguistics~\citep{Edmonds-Wathen852956, VUKOVIC2015110}.
\citet{Levinson_2003} initially defines three FoR classes: \textit{relative}, based on the observer’s perspective; \textit{intrinsic}, based on an inherent feature of the reference object; and \textit{absolute}, using environmental cues like cardinal directions (see Figure~\ref{fig:FoR_classes}).
This framework was expanded by~\citet{TENBRINK2011704} to a more comprehensive framework, serving as the basis of this paper.
Understanding FoR is important for many applications, especially in embodied AI.
In such applications, an agent must simultaneously comprehend multiple perspectives, including the one from the instruction giver and from the instruction follower, to communicate and perform tasks effectively~\cite{liu-etal-2013-modeling}.
However, recent spatial evaluation benchmarks have largely overlooked FoR.
For example, the text-based benchmarks~\citet{mirzaee-etal-2021-spartqa, shi2022stepgamenewbenchmarkrobust, mirzaee2022transferlearningsyntheticcorpora, rizvi2024sparcsparpspatialreasoning} and text-to-images benchmarks~\citep{{gokhale2023benchmarkingspatialrelationshipstexttoimage, huang2023t2icompbenchcomprehensivebenchmarkopenworld, cho2023dallevalprobingreasoningskills, cho2023visualprogrammingtexttoimagegeneration}}  assume a fixed perspective for all spatial expressions. 
This inherent bias limits situated spatial reasoning, restricting adaptability in interactive environments where perspectives can change.

To systematically investigate the role of FoR in spatial understanding, we create a new resource, \textbf{F}rame \textbf{o}f \textbf{R}eference \textbf{E}valuation in \textbf{S}patial Reasoning \textbf{T}asks (FoREST),
to evaluate models’ ability to comprehend FoR from textual descriptions and extend this evaluation to grounding and visualization. Our benchmark includes spatial expressions with FoR ambiguity—cases where multiple FoRs may apply to the described situation—as well as spatial expressions with only a single valid FoR. 
This design allows evaluation of the models' understanding of FoR in both scenarios. We evaluate several LLMs in a QA setting that requires FoR understanding and employ the FoR concept in text-to-image models. 
Our findings reveal performance differences across FoR classes and show that LLMs exhibit bias toward specific FoRs when handling ambiguous cases.
This bias extends to layout-diffusion models, which rely on LLM-generated layouts in the image generation pipeline.
To enhance FoR comprehension in LLMs, we propose Spatial-Guided prompting, which enables models to analyze and extract additional spatial information, including directional, topological, and distance relations. 
We demonstrate that incorporating spatial primitives and relations improves question-answering and layout generation, ultimately enhancing text-to-image generation performance.

Our contribution\footnote{code and dataset available at
~\href{https://github.com/HLR/FoREST}{Github repository}.} 
are summarized as follows, 
1) We introduce the FoREST benchmark to systematically evaluate LLMs’ FoR comprehension, 
2) We analyze the impact of FoR information on text-to-image generation using diffusion models,
3) We propose a prompting approach that generates spatial primitives and relations in the chain of reasoning, which enhances the performance of QA and layout diffusion models.


\section{Spatial Primitives}\label{sec:primitives}
We review three semantic aspects of spatial information expressed in language: Spatial Roles, Spatial Relations, and Frame of Reference.  

\noindent\textbf{Spatial Roles.} 
We focus on two main spatial roles~\citep{kordjamshidi-etal-2010-spatial} of \textit{Locatum}, and \textit{Relatum}. 
The locatum is the object described in the spatial expression, while the relatum is the other object used to describe the position of the locatum. 
An example is \textit{a cat is to the left of a dog}, where the \textit{cat} is the locatum, and the \textit{dog} is the relatum.

\noindent\textbf{Spatial Relations.} 
When dealing with spatial knowledge representation and reasoning, three main relations are often considered: directional, topological, and distal~\citep{reasoningQualitaiveDaniel, COHN2008551, ACMpaper}. 
\textit{Directional} describes an object's direction based on specific coordinates, e.g., left and right.
\textit{Topological} describes the containment between two objects, such as inside.
\textit{Distal} describes qualitative and quantitative relations regarding the distance between entities. An example of a qualitative distal relation is \textit{far}, and an example of a quantitative one is $3\text{km}$.

\noindent\textbf{Frame of Reference.} We use four frames of reference investigated in the cognitive linguistic studies~\cite{TENBRINK2011704}. These are defined based on the concept of \textit{Perspective}, which is the origin of a coordinate system to determine the direction. The four frames of reference are defined as follows.

\noindent1. \textit{External Intrinsic} describes a spatial relation from the relatum's perspective, where the relatum does not contain the locatum. The top-right image in Figure~\ref{fig:FoR_classes} shows this case with the sentence, \textit{A cat is to the right of the car from the car's perspective.}

\noindent2. \textit{External Relative} describes a spatial relation from the observer's perspective.
The top-left image in Figure~\ref{fig:FoR_classes}  shows an example with the sentence, \textit{A cat is to the left of a car from my perspective.}

\noindent3. \textit{Internal Intrinsic} describes a spatial relation from the relatum's perspective, where the relatum contains the locatum. The bottom-right of Figure~\ref{fig:FoR_classes} shows this case with the sentence, \textit{A cat is inside and back of the car from the car's perspective.}

\noindent4. \textit{Internal Relative} describes a spatial relation from the observer's perspective where the locatum is inside the relatum. The bottom-left image in Figure~\ref{fig:FoR_classes} shows this case with the sentence, \textit{A cat is inside and to the left of the car from my perspective.}

\begin{figure*}[t!]
    \centering
    \includegraphics[width=0.87\linewidth]{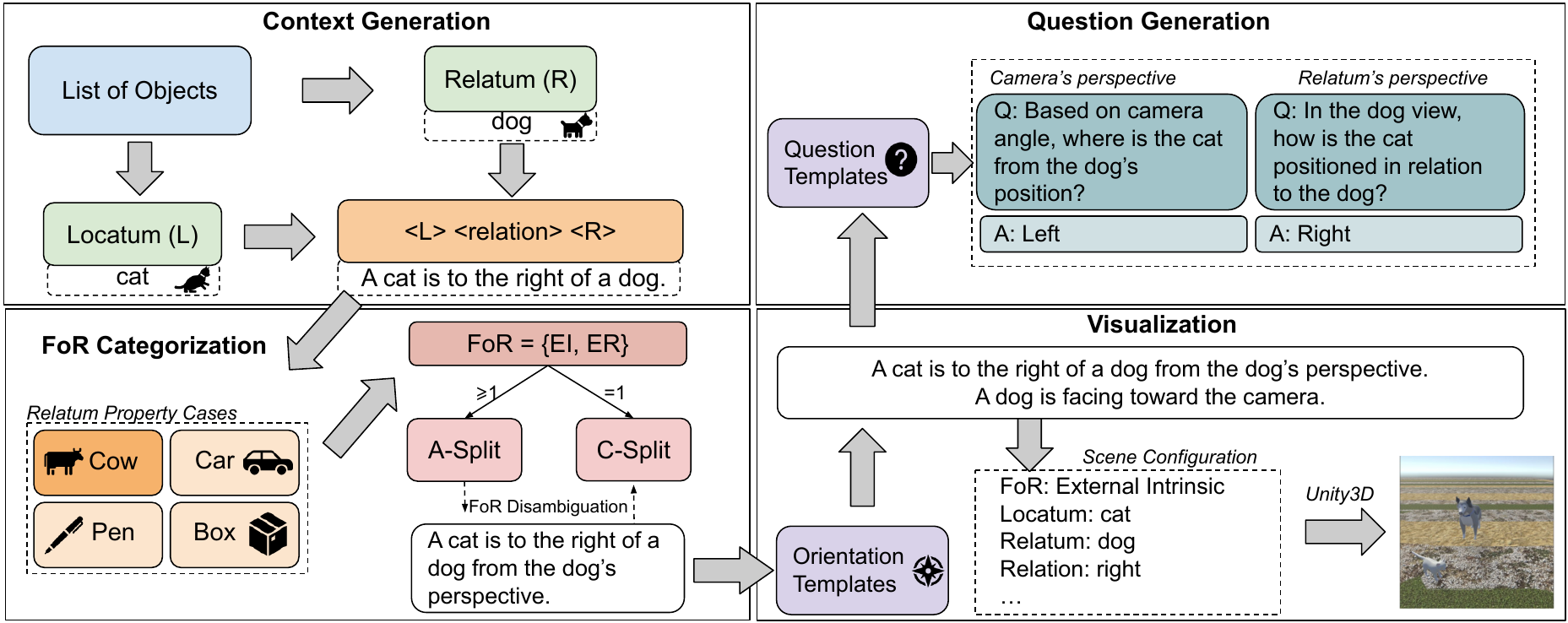}
    \caption{The dataset creation pipeline. It begins by selecting a locatum and a relatum from a pre-defined list of objects and then applies templates to generate the spatial expressions ($T$). FoRs are then assigned based on the relatum properties. $T$ is categorized based on the number of applicable FoRs. For example, \textit{A cat is to the right of a dog} (with two possible FoRs: external intrinsic and external relative) belongs to the A-split. Then, its disambiguated version (A cat is to the right of a dog from the dog's perspective) is added to the C-split. Next, if applicable, the relatum orientation is included for visualization and question generation. Finally, Unity3D generates the scene configurations, and the question-answer pairs are derived from $T$.}
    \label{fig:generate_pipeline_image}
\end{figure*}

\section{FoREST Dataset Construction}\label{sec:DatasetCreation}


To systematically evaluate LLMs on the frame of reference (FoR) recognition, 
we introduce the \textbf{F}rame \textbf{o}f \textbf{R}eference \textbf{E}valuation in \textbf{S}patial Reasoning \textbf{T}asks (FoREST) benchmark.
Each instance in FoREST consists of a spatial context ($T$), a set of corresponding FoRs ($FoR$) which is a subset of \{\textit{external relative},  \textit{external intrinsic}, \textit{internal intrinsic}, \textit{internal relative}\}, a set of questions and answers ($\{Q,A\}$), and a set of visualizations ($\{I\}$).
An example of $T$ is \textit{``A cat is to the right of a dog. A dog is facing toward the camera.''}
The set of applicable FoRs for $T$ is \{\textit{external intrinsic}, \textit{external relative}\}.
A question-answer pair is $Q$ = \textit{``Based on the camera's perspective, where is the cat from the dog's position?''}, $A$ = \{left, right\}. There is an ambiguity in the FoR for this expression.
Thus, the answer will be \textit{left} if the model assumes the external relative FoR. In contrast, it will be \textit{right} if the model assumes the external intrinsic FoR.
The visualization of this example is shown in Figure~\ref{fig:generate_pipeline_image}. 
Dataset statistics are provided in Table~\ref{tab:data_statistic}.

\subsection{Context Generation}
We select two distinct objects—a relatum ($R$) and a locatum ($L$)—from a set of 20 objects and apply them to a Spatial Relation template,
\textit{<$L$> <spatial relation> <$R$>}, to generate the context $T$.
FoRs for $T$ are determined based on the properties of the selected objects. Depending on the number of possible FoRs, $T$ is categorized as ambiguous (A-split), where multiple FoRs apply, or clear (C-split), where only one FoR is valid. 
We further augment the C-split with disambiguated spatial expressions derived from the A-split, as shown in Figure~\ref{fig:generate_pipeline_image}.

\subsection{Categories based on Relatum Properties} \label{sec:FoR_Relatum_scenario}
Using the FoR classes in Section~\ref{sec:primitives}, we identify two key properties contributing to FoR ambiguity.
The first property is the relatum's intrinsic direction. It creates ambiguity between intrinsic and relative FoRs, since spatial relations may originate from either the relatum’s or the observer’s perspective. 
The second property is the relatum's affordance as a container. 
It introduces the ambiguity between internal and external FoR, as spatial relations may refer to either inside or outside of the relatum. 
{Note that containment is defined as the ability of the relatum to contain the locatum, considering both objects’ sizes.}
Based on these properties, we define four distinct cases: \textit{Cow, Box, Car, and Pen.}

\noindent\textbf{Case 1: Cow Case}.
In this case, the selected relatum has intrinsic directions but cannot be the container for the locatum. An example object is a cow.
In such cases, the relatum provides a perspective for spatial relations.
The applicable FoR classes are $FoR$ = \{\textit{external intrinsic}, \textit{external relative}\}.
We augment the C-split with expressions of this case, but include the perspective to resolve their ambiguity.
To specify the perspective, we use predefined templates for augmenting clauses, such as \textit{from \{relatum\}'s perspective }for \textit{external intrinsic} or \textit{from the camera's perspective} for \textit{external relative}. 
For example, if the context is  \textit{``A cat is to the left of the cow''}, in the A-split,
the counterparts included in the C-split are \textit{``A cat is to the left of the cow from the cow's perspective.''} for the \textit{external intrinsic} and \textit{``A cat is to the left of the cow from my perspective.''} for the \textit{external intrinsic}.

\noindent\textbf{Case 2: Box Case.} 
In this case, the relatum has the container affordance but lacks intrinsic directions, e.g, a box. 
The applicable FoR classes are $FoR$ = \{\textit{external relative}, \textit{internal relative}\}.
To include unambiguous counterparts in the C-split, we specify a topological relation by adding \textit{inside} for \textit{internal relative} and \textit{outside} for \textit{external relative}. 
For example, given the sentence \textit{``A cat is to the left of the box.''},
the unambiguous $T$ with \textit{internal relative} FoR is \textit{``A cat is inside and to the left of the box.''} The counterpart for \textit{external relative} is \textit{``A cat is outside and to the left of the box.''}

\noindent\textbf{Case 3: Car Case.}  
A relatum with an intrinsic direction and container affordance falls into this case. An example object is a car. The applicable FoR classes are $FoR$ = \{ \textit{external relative},  \textit{external intrinsic}, \textit{internal intrinsic}, \textit{internal relative}\}.
To augment C-split with the disambiguated counterparts of such cases, we add perspective and topology information to the sentences similar to the Cow and Box cases.
An example expression for this case is \textit{A person is in front of the car.} 
The four disambiguated counterparts to include in the C-split are \textit{``A person is outside and in front of the car from the car itself.''} for \textit{external intrinsic}, \textit{``A person is outside and in front of the car from the observer.''} for \textit{external relative},  \textit{``A person is inside and in front of the car from the car itself.''} for \textit{internal intrinsic}, and \textit{``A person is inside and in front of the car from the observer.''} for \textit{internal relative}.

\noindent\textbf{Case 4: Pen Case.} 
In this case, the relatum lacks both the intrinsic direction and the affordance as a container. An example object is a pen.
Lacking these two properties, the created context has only one applicable FoR, $FoR$ = \{\textit{external relative}\}.
Therefore, we can categorize this case into both splits without any modification.
An example of such a context is \textit{``The book is to the left of a pen.''}

\subsection{Context Visualization}\label{sec:context_visualize}
In our visualization, complexity arises when the relatum has an intrinsic direction, as its orientation can complicate the spatial representation.
For example, for the visualization of \textit{A cat is to the right of a dog from the dog's view}, the cat can be placed in different coordinates based on the dog’s orientation.
To address this issue, we add a template sentence for each direction, such as \textit{<relatum> is facing toward the camera}, to specify the relatum's orientation of all applicable $T$ for visualization and QA.
For instance, \textit{``A cat is to the left of a dog.''} becomes \textit{``A cat is to the left of a dog. The dog is facing toward the camera.''}.
To avoid occlusion, we generate visualizations only for external FoRs, since one object may become invisible in internal FoR classes. 
We use only C-split expressions, as they have a unique FoR interpretation for visualization. A scene configuration is then created by applying a predefined template, as illustrated in Figure~\ref{fig:generate_pipeline_image}.
The images are generated using the Unity 3D simulator~\cite{juliani2020unitygeneralplatformintelligent}, producing four variations per expression $T$ with different backgrounds and object positions. 
Further details of the creation process are provided in Appendix~\ref{appendix:dataset_creation}.

\subsection{Question-Answering Generation}\label{sec:QA_generation}
We generate corresponding questions for each spatial expression ($T$). When the relatum has an intrinsic direction, we also include its orientation as described in Section~\ref{sec:context_visualize}. 
Our benchmark contains two question types. 
The first asks for the spatial relation between two objects from the camera’s perspective, following predefined templates such as, \textit{Based on the camera’s perspective, where is the {locatum} relative to the {relatum}’s position?}
The second queries the relation from the relatum’s perspective, using the same templates but replacing the camera with the relatum.
The first type is generated for all $T$, while the second applies only when the relatum has an intrinsic direction. 
Answers are determined by the corresponding FoR, the spatial relation expressed in $T$, and the relatum’s orientation when applicable.
Question template variations were generated using GPT-4o, with details provided in Appendix~\ref{appendix:textual_template}.
To demonstrate that the order of the locatum and the relatum in the question does not affect LLMs' performance, we also evaluate Qwen2-72B on templates with the reversed order of the locatum and the relatum in Appendix~\ref{appendix:inverted_question}. Qwen2-72B performs similarly on both orders, so we use only the question templates where the locatum precedes the relatum in the remaining experiments.

\begin{table}[t]
    \centering
    \scriptsize
    \begin{tabular}{c c c c c}
        \toprule
        \multicolumn{5}{c}{A-split} \\
        \hline
         Case & Context	& QA-CP	& QA-RP& T2I \\
         \hline
         Cow Case & 792 & 3168 & 3168 & 3168 \\
         Car Case & 128 & 512 & 512 & 512 \\
         Box Case & 120 & 120 & 120 & 120 \\
         Pen Case & 488 & 488 & 488 & 488 \\
         \hline
         Total &1528 & 4288 & 4288 & 4288 \\
         \hline
         \multicolumn{5}{c}{C-Split} \\
        \hline
        FoR Class & Context& QA-CP	& QA-RP & T2I \\
         \hline
         External Relative & 1528 & 4288 & 3680 & 4288 \\
         External Intrinsic & 920 & 3680 & 3680 & 3680 \\
         Internal Intrinsic & 128 & 512 & 512 & 0 \\
         Internal Relative & 248 & 632 & 512 & 0 \\
         \hline
         Total & 2824 & 9112 & 8384 & 7968 \\
         \bottomrule
    \end{tabular}
    \caption{Dataset statistics of FoREST A-split and C-split portions. QA-CP and QA-RP are question-answer pairs with camera perspective and question-answer pairs with relatum perspective, respectively. T2I refers to the prompt used in Text-to-Image experiments.}
    \label{tab:data_statistic}
\end{table}

\section{Models and Tasks} 
This paper focuses on Question-Answering and Text-to-Image tasks using the FoREST benchmark to evaluate FoR in spatial reasoning comprehensively. 
FoREST supports additional tasks, such as FoR identification, detailed in Appendix~\ref{appendix:FoRIdentification}.

\subsection{Question-Answering (QA)}\label{sec:QA_explanation}

\noindent\textbf{Task.}
The QA task evaluates LLMs’ ability to adapt contextual perspectives across different FoRs.
The input includes a spatial expression $T$, relatum orientation (if available), and a question $Q$ querying the spatial relation from either the observer's or the relatum’s perspective. 
The output is a spatial relation $S$, restricted to \{left, right, front, back\}.
We use both dataset splits for this task.

\noindent\textbf{Zero-shot baseline.} 
We call the LLM with instructions, a spatial context, and a question expecting a spatial relation as the response. 
The prompt instructs the model to answer the question with one of the candidate spatial relations without explanation.

\noindent\textbf{Few-shot baseline.} 
We create four spatial expressions, each assigned to a single FoR class to prevent bias. 
Following the steps in Section~\ref{sec:QA_generation}, we generate a corresponding question and answer for each. These serve as examples in our few-shot prompting. The input to the model is the instruction, example, spatial context, and the question.

\noindent\textbf{Chain-of-Thought (CoT) baseline~\citep{wei2023chainofthoughtpromptingelicitsreasoning}.}
To create CoT examples, we modify the prompt to require reasoning before answering.
We manually craft reasoning explanations with the necessary information for the few-shot examples.
The input to the model is the instruction, CoT example, spatial context, and the question.

\noindent\textbf{Human baseline.}
We conducted a small-scale human study with three participants, compensated as research assistants. Each participant was shown 25 randomly selected QA examples per relatum case in the A-split, for a total of 150 examples. Each example consisted of a textual scene description and a corresponding question.
See Appendix~\ref{appendix:human_study} for details. We report average accuracy across participants to reflect the human baseline.

\begin{table*}[t]
    \setlength{\tabcolsep}{0.9mm}
    \small
    \centering
    \begin{tabular}{l c c c | c c c | c | c | c | c c c | c c c | c }
    \toprule
     & \multicolumn{9}{c |}{Question with Camera Perspective} & \multicolumn{7}{c}{Question with Relatum Perspective} \\
    \cline{2-17}
     Model & \multicolumn{3}{c|}{Cow} & \multicolumn{3}{c | }{Car} & Box & Pen & Avg. &\multicolumn{3}{c |}{Cow} & \multicolumn{3}{c |}{Car} & Avg. \\
      \cline{2-17}
       & R\% & I\% & Acc. & R\% & I\% & Acc. & Acc. & Acc. & Acc. &R\% & I\% & Acc. & R\% & I\% & Acc. & Acc. \\ 
        \hline
Llama3-70B (1) & $48.1$ & $\mathbf{51.5}$ & $62.5$ & $\mathbf{58.0}$ & $41.6$ & $65.5$ & $73.3$ & $72.5$ & $64.3$  & $\mathbf{61.0}$ & $38.7$ & $62.1$ & $\mathbf{51.8}$ & $47.9$ & $61.8$ & $62.1$ \\
Llama3-70B (2) & $49.1$ & $\mathbf{50.5}$ & $62.2$ & $\mathbf{52.2}$ & $47.4$ & $64.7$ & $85.8$ & $85.5$ & $65.8$  & $\mathbf{59.6}$ & $40.1$ & $57.1$ & $\mathbf{55.5}$ & $44.2$ & $61.8$ & $57.7$ \\
Llama3-70B (3) & $49.4$ & $\mathbf{50.3}$ & $80.7$ & $49.4$ & $\mathbf{50.3}$ & $79.6$ & $95.8$ & $94.9$ & $82.6$  & $\mathbf{60.8}$ & $39.0$ & $77.2$ & $\mathbf{55.1}$ & $44.6$ & $80.9$ & $77.7$ \\
Llama3-70B (4) & $\mathbf{59.4}$ & $40.2$ & $73.6$ & $\mathbf{57.9}$ & $41.7$ & $74.8$ & $100.0$ & $100.0$ & $77.5$  & $\mathbf{60.6}$ & $39.1$ & $65.7$ & $\mathbf{56.0}$ & $43.7$ & $67.7$ & $66.0$ \\
\hline
Qwen2-72B (1) & $\mathbf{96.6}$ & $2.9$ & $95.6$ & $\mathbf{95.9}$ & $3.6$ & $95.0$ & $100.0$ & $100.0$ & $96.1$  & $8.8$ & $\mathbf{90.6}$ & $79.3$ & $7.8$ & $\mathbf{91.7}$ & $83.6$ & $79.9$ \\
Qwen2-72B (2) & $\mathbf{89.0}$ & $10.5$ & $84.4$ & $\mathbf{85.6}$ & $13.9$ & $85.5$ & $100.0$ & $100.0$ & $86.8$  & $17.7$ & $\mathbf{81.8}$ & $78.3$ & $10.4$ & $\mathbf{89.1}$ & $86.3$ & $79.4$ \\
Qwen2-72B (3) & $\mathbf{67.2}$ & $32.4$ & $88.6$ & $\mathbf{62.0}$ & $37.6$ & $83.4$ & $100.0$ & $100.0$ & $89.6$  & $21.3$ & $\mathbf{78.3}$ & $85.5$ & $22.7$ & $\mathbf{76.9}$ & $83.6$ & $85.2$ \\
Qwen2-72B (4) & $\mathbf{93.0}$ & $6.5$ & $90.1$ & $\mathbf{94.6}$ & $4.9$ & $93.3$ & $100.0$ & $98.6$ & $91.7$  & $8.2$ & $\mathbf{91.2}$ & $86.0$ & $10.5$ & $\mathbf{89.0}$ & $87.4$ & $86.2$ \\
\hline
Qwen2VL-72B (1) & $49.5$ & $\mathbf{50.5}$ & $78.3$ & $\mathbf{51.6}$ & $48.4$ & $80.0$ & $98.3$ & $96.9$ & $81.2$  & $41.0$ & $\mathbf{59.0}$ & $55.4$ & $44.7$ & $\mathbf{55.3}$ & $59.0$ & $56.0$ \\
Qwen2VL-72B (2) & $40.9$ & $\mathbf{59.1}$ & $89.4$ & $44.7$ & $\mathbf{55.3}$ & $79.0$ & $100.0$ & $100.0$ & $89.6$  & $32.7$ & $\mathbf{67.3}$ & $66.9$ & $28.5$ & $\mathbf{71.5}$ & $67.0$ & $66.9$ \\
Qwen2VL-72B (3) & $\mathbf{63.5}$ & $36.5$ & $84.0$ & $\mathbf{72.2}$ & $27.8$ & $84.5$ & $100.0$ & $100.0$ & $86.3$  & $\mathbf{51.0}$ & $49.0$ & $77.7$ & $\mathbf{54.6}$ & $45.4$ & $82.8$ & $78.4$ \\
Qwen2VL-72B (4) & $\mathbf{50.5}$ & $49.5$ & $78.3$ & $\mathbf{59.5}$ & $40.5$ & $67.4$ & $98.3$ & $99.6$ & $79.9$  & $23.5$ & $\mathbf{76.5}$ & $78.2$ & $33.0$ & $\mathbf{67.0}$ & $67.7$ & $76.7$ \\
\hline
GPT-4o (1) & $\mathbf{84.3}$ & $15.3$ & $94.5$ & $\mathbf{88.5}$ & $11.0$ & $97.3$ & $99.2$ & $99.8$ & $95.6$  & $21.6$ & $\mathbf{78.0}$ & $91.6$ & $16.1$ & $\mathbf{83.5}$ & $90.5$ & $91.4$ \\
GPT-4o (2) & $\mathbf{69.0}$ & $30.6$ & $76.6$ & $\mathbf{80.3}$ & $19.2$ & $89.5$ & $100.0$ & $100.0$ & $81.5$  & $29.0$ & $\mathbf{70.5}$ & $74.7$ & $30.9$ & $\mathbf{68.7}$ & $77.5$ & $75.1$ \\
GPT-4o (3) & $41.5$ & $\mathbf{58.3}$ & $92.3$ & $38.2$ & $\mathbf{61.6}$ & $91.0$ & $100.0$ & $99.8$ & $93.2$  & $33.9$ & $\mathbf{65.8}$ & $93.9$ & $32.0$ & $\mathbf{67.6}$ & $93.9$ & $93.9$ \\
GPT-4o (4) & $26.0$ & $\mathbf{73.9}$ & $79.2$ & $27.7$ & $\mathbf{72.1}$ & $79.4$ & $96.7$ & $94.3$ & $81.4$  & $16.2$ & $\mathbf{83.4}$ & $95.5$ & $19.2$ & $\mathbf{80.4}$ & $94.8$ & $95.4$ \\
\hline
Human-baseline & $36.6$ & $\mathbf{63.4}$ & $90.7$ & $27.8$ & $\mathbf{72.2}$ & $96.0$ & $72.0$ & $82.7$ & $85.3$ & $41.4$ & $\mathbf{58.6}$ & $97.3$ & $36.1$ & $\mathbf{63.9}$ & $96.0$ & $96.7$ \\
\bottomrule
    \end{tabular}
    \caption{QA accuracy in the A-split. R\% and I\% indicate the proportion of cases where the model assumes a relative or intrinsic FoR for an ambiguous expression (see Section~\ref{sec:evaluation_setting}). Acc denotes accuracy, and Avg is the micro-average accuracy. (1) 0-shot, (2) 4-shot, (3) CoT, and (4) SG+CoT. }
    \label{tab:A_split-QA}
\end{table*}

\subsection{Text-To-Image (T2I)}\label{sec:t2i_models}

\noindent\textbf{Task.} This task assesses the ability of diffusion models to consider FoR by evaluating their generated images. The input is a spatial expression, $T$, and the output is a generated image ($I$). C and A splits with external FoRs are used for this task.

\noindent\textbf{Stable Diffusion Model.} 
We use the stable diffusion model as the baseline for the T2I task. 
This model only needs the scene description as input. 

\noindent\textbf{Layout Diffusion Model.}
This model operates in two phases: text-to-layout and layout-to-image.
Given that the LLM can generate the bounding box layout~\citep{cho2023visualprogrammingtexttoimagegeneration}, we provide the LLM with instructions and $T$ to create the layout. 
The layout consists of bounding box coordinates for each object in the format of \{object: $[x, y, w, h]$\}, where $x$ and $y$ denote the starting point and $h$ and $w$ denote the height and width. 
The bounding box coordinates and $T$ are then passed to the layout-to-image model to produce the final image, $I$. 

\subsection{Spatial-Guide (SG) Prompting}\label{sec:SG_prompting}
We hypothesize that the spatial relation types and FoR classes defined in Section~\ref{sec:primitives} can guide QA and layout generation.
For example, the \textit{external intrinsic} FoR emphasizes that spatial relations originate from the relatum’s perspective.
To leverage this, we propose SG prompting, an additional step applied before QA or layout generation.
This step extracts spatial information, including direction, topology, distance, and the FoR, from the spatial expression $T$, as supplementary input to guide LLMs in QA or layout generation.
We manually craft four examples covering these aspects.
First, we specify the perspective for \textit{directional relations}, e.g., \textit{left} relative to the observer, to distinguish intrinsic from relative FoR.
Next, we indicate whether the locatum is inside or outside the relatum for \textit{topological relations} to differentiate internal from external FoR.
Lastly, we provide an estimated quantitative distance to support topological and directional relation identification, e.g., \textit{far}.
These examples are provided in a few-shot setting to guide the model in automatically extracting such information. 
{The extracted information is then used to guide CoT reasoning in QA and layout generation.}

\section{Experimental Results}
\begin{table*}[ht!]
    \small
    \setlength{\tabcolsep}{0.9mm}
    \centering
    \begin{tabular}{l c c c c c |c c c c c}
    \toprule
     & \multicolumn{5}{c|}{Question with Camera Perspective} & \multicolumn{5}{c}{Question with Relatum Perspective} \\
    \cline{2 - 11}
    Model & ER (CP) & EI (RP) & II (RP) & IR (CP) & Avg. & ER (CP) & EI (RP) & II (RP) & IR (CP) & Avg. \\
    \hline
     Llama3-70B (0-shot) & $44.8$ & $38.4$ & $39.7$ & $54.4$ & $42.6$ &$42.2$ & $47.1$ & $62.5$ & $34.4$ & $45.1$ \\
     Llama3-70B (4-shot) & $43.0$ & $40.0$ & $39.1$ & $47.3$ & $41.9$ & $41.8$ & $60.9$ & $77.7$ & $35.2$ & $52.0$ \\
     Llama3-70B (CoT) & $57.8$ & $46.1$ & $44.7$ & $46.0$ & $51.5$ & $\mathbf{55.5}$ & $56.8$ & $71.5$ & $49.0$ & $56.6$ \\
     Llama3-70B (SG+CoT) & $47.6$ & $42.9$ & $50.0$ & $35.6$ & $45.0$ &$55.4$ & $64.5$ & $75.0$ & $47.1$ & $60.1$ \\
     \hline
     Qwen2-72B (0-shot) & $94.5$ & $35.2$ & $31.8$ & $93.2$ & $66.9$ & $28.7$ & $89.3$ & $93.6$ & $23.8$ & $59.0$ \\
     Qwen2-72B (4-shot) & $90.2$ & $39.5$ & $39.1$ & $68.5$ & $65.3$ & $33.5$ & $92.1$ & $94.0$ & $29.5$ & $62.7$ \\
     Qwen2-72B (CoT) & $81.4$ & $57.4$ & $58.6$ & $62.5$ & $69.1$ & $39.5$ & $83.7$ & $85.2$ & $37.7$ & $61.6$ \\
     Qwen2-72B (SG+CoT) & $97.6$ & $42.5$ & $31.3$ & $\mathbf{93.8}$ & $71.4$ & $42.8$ & $86.6$ & $92.0$ & $34.0$ & $64.5$ \\
     \hline
    Qwen2VL-72B (0-shot)   & $68.3$ & $40.2$ & $42.4$ & $75.3$ & $56.0$ & $37.4$ & $60.1$ & $82.2$ & $35.5$ & $50.0$ \\
    Qwen2VL-72B (4-shot)   & $78.4$ & $41.0$ & $44.5$ & $66.1$ & $60.5$ & $38.0$ & $85.4$ & $91.2$ & $29.9$ & $61.6$ \\
    Qwen2VL-72B (CoT)   & $58.4$ & $62.4$ & $68.4$ & $40.7$ & $59.3$ & $48.6$ & $43.5$ & $36.1$ & $\mathbf{61.9}$ & $46.4$ \\
    Qwen2VL-72B (SG+CoT) & $\mathbf{99.0}$ & $50.0$ & $47.5$ & $93.0$ & $75.9$ & $27.3$ & $37.3$ & $37.1$ & $34.0$ & $32.7$ \\
     \hline
    GPT-4o (0-shot)  & $79.7$ & $45.1$ & $39.5$ & $90.2$ & $64.2$  & $46.9$ & $88.5$ & $98.2$ & $34.8$ & $67.5$ \\
    GPT-4o (4-shot) & $68.0$ & $52.6$ & $60.7$ & $74.1$ & $61.8$  & $44.9$ & $\mathbf{98.2}$ & $\mathbf{100.0}$ & $37.5$ & $71.2$ \\
    GPT-4o (CoT) & $81.7$ & $\mathbf{76.1}$ & $\mathbf{82.4}$ & $71.5$ & $78.8$  & $53.0$ & $91.1$ & $90.6$ & $50.8$ & $\mathbf{71.9}$ \\
    GPT-4o (SG+CoT)  & $97.9$ & $72.2$ & $72.7$ & $93.4$ & $\mathbf{85.8}$  & $48.9$ & $96.3$ & $95.9$ & $36.1$ & $71.8$ \\
    \bottomrule
    \end{tabular}
    \caption{QA accuracy in the C-Split across various LLMs. ER, EI, II, and IR denote external relative, external intrinsic, internal intrinsic, and internal relative FoRs, respectively. Avg denotes the micro-average accuracy. CP indicates context with a camera perspective, while RP denotes context with a relatum perspective.}
    \label{tab:QA_c_split}
\end{table*}

\begin{table*}[t]
    \centering
    \setlength{\tabcolsep}{0.9mm}
    \small
    \begin{tabular}{l  c c c | c  c | c  c  c  c }
    \toprule
         & \multicolumn{8}{c}{VISOR(\%)} \\ \cline{2-9}
          & \multicolumn{5}{c|}{ A-Split } &  \multicolumn{3}{c}{ C-Split }\\ \cline{2-5} \cline{6 - 9}
        Model & \multicolumn{3}{c |}{cond (I)} & cond (R) & cond (avg) & cond (I) & cond (R) & cond (avg) \\ \cline{2-4}
        & EI FoR & ER FoR & all & & & & & \\ \hline
        SD-1.5   & $ 51.11$  & $ 21.61$  &  $ 72.72$ & $ 48.95$ & $ 68.72$ & $ 53.92$ & $ 53.77$ & $ 53.83$ \\
        SD-2.1  & $ 57.97$  & $ 21.49$ &  $ 79.46$ & $ 54.10$ & $ 75.39$ & $\mathbf{60.06}$ & $ 59.64$ & $ 59.83$ \\
        \hline
        Llama3-8B + GLIGEN& $ 53.67$  & $ 25.78$ & $ 79.45$ & $ 66.08$ & $ 77.38$ & $ 57.51$ & $ 65.98$ & $ 62.12$ \\
        Llama3-70B + GLIGEN & $ 54.49$  & $ 29.45$ & $ 83.94$ & $ 68.68$ & $ 81.43$ & $ 56.47$ & $ 69.53$ & $ 63.49$ \\
        Llama3-8B + SG + GLIGEN (Our) & $ 57.46$  & $ 27.96$ & $ 85.42$ & $\mathbf{71.14}$ & $ 83.17$ & $ 58.84$ & $\mathbf{70.36}$ & $ \mathbf{65.15}$ \\
        Llama3-70B + SG + GLIGEN (Our)  & $ 56.54$  & $ 30.59$ & $ \mathbf{87.13}$ & $ 66.56$ & $\mathbf{83.75}$ & $ 56.77$ & ${70.04}$ & ${64.06}$ \\
        \bottomrule
    \end{tabular}
    \caption{VISOR$_{cond}$ score explained in Section~\ref{sec:evaluation_setting} where $I$ refers to the Cow and Car cases where relatum has intrinsic directions, and $R$ refers to the Box and Pen cases where relatum lacks intrinsic directions, $avg$ is the micro-average of $I$ and $R$. EI and ER FoR represent the generated image considered corrected by EI or ER FoR. }
    \label{tab:I_split}
\end{table*}

\subsection{Evaluation Metrics}\label{sec:evaluation_setting}
\noindent\textbf{QA.} We report accuracy (acc.) based on the correct answer defined as follows. Since the questions can have multiple correct answers, as explained in Section~\ref{sec:DatasetCreation}, the prediction is correct if it matches any valid answer.
Additionally, we report the model’s bias distribution when FoR ambiguity exists.
$I$\% is the percentage of correct answers when assuming an intrinsic FoR, while $R$\% is this percentage with a relative FoR assumption. 
Note that cases where both FoR assumptions lead to the same answer are excluded from the bias calculation.

\noindent\textbf{T2I.} 
We adopt \textit{spatialEval}~\citep{cho2023visualprogrammingtexttoimagegeneration} for evaluating T2I spatial ability. 
We modify it to account for FoR by converting relations to a camera perspective before passing them to spatialEval, which assumes this viewpoint.
Accuracy is determined by comparing the bounding box and depth map of the relatum and locatum. 
For FoR ambiguity, a generated image is correct if it aligns with at least one valid FoR interpretation.
We report results using VISOR$_{cond}$ and VISOR$_{uncond}$~\citep{gokhale2023benchmarkingspatialrelationshipstexttoimage} metrics. 
VISOR$_{cond}$ evaluates spatial relations only when both objects appear correctly, aligning with our focus on spatial reasoning. While, VISOR$_{uncond}$ evaluates the overall performance, including object creation errors.

\subsection{Experimental Setting}
\noindent\textbf{QA.} 
We use Llama3-70B~\citep{dubey2024Llama3herdmodels}, Qwen2-72B~\citep{qwen2model}, Qwen2VL-72B~\cite{wang2024qwen2vlenhancingvisionlanguagemodels}, and GPT-4o~\citep{openai2024gpt4technicalreport} as backbone models. All models are evaluated with \textit{zero-shot}, \textit{few-shot}, \textit{CoT}, and our SG+CoT prompting under temperature 0 to ensure reproducibility. 

\noindent\textbf{T2I.}
We select Stable Diffusion SD-1.5 and SD-2.1~\citep{rombach2021highresolution} as our stable diffusion models and GLIGEN\citep{li2023gligenopensetgroundedtexttoimage} as the layout-to-image backbone.
For translating spatial descriptions into bounding box information, we use Llama3-8B and Llama3-70B, as detailed in Section~\ref{sec:t2i_models}. 
The same LLMs are used to generate spatial information for SG prompting. 
We generate four images to compute the VISOR score following~\cite{gokhale2023benchmarkingspatialrelationshipstexttoimage}
Inference steps for all T2I models are set to 50.
For the evaluation, we select grounding DINO~\citep{liu2024groundingdinomarryingdino} for object detection and DPT~\citep{ranftl2021visiontransformersdenseprediction} for depth mapping, following VPEval~\cite{cho2023visualprogrammingtexttoimagegeneration}. The experiments were conducted on two $A6000$ GPUs, totaling approximately $300$ GPU hours.

\subsection{Results}
\noindent\textbf{RQ1. What is the LLM's bias when FoR is ambiguous? }
Table~\ref{tab:A_split-QA} presents the QA results for the A-split. 
Since the context lacks a fixed perspective, a model extracting spatial relations alone should ideally reach 100\% accuracy. Our goal, however, is to assess LLMs’ bias by measuring how often they adopt a specific perspective. In the Box and Pen cases, relatum properties do not introduce FoR ambiguity in directional relations, making the task pure extraction rather than reasoning.
Thus, we focus on the $I$\% and $R$\% of the Cow and Car cases, which best reflect LLMs’ bias.
Qwen2-72B consistently achieves 80–95\% accuracy across all experiments by selecting spatial relations directly from context, suggesting it may disregard the question’s perspective.
This is supported by attention analysis in Appendix~\ref{appendix:attention}. The attention maps show that most correct Qwen2 responses assume a shared perspective between context and question and pay low attention to the perspective stated in the question.
GPT-4o exhibits similar bias in 0-shot and 4-shot settings but shifts toward intrinsic interpretation with CoT. This shift reduces accuracy on camera-perspective questions, where FoR adaptation plays a larger role than relation extraction. Llama3-70B shows no strong preference, slightly favoring relative FoR, but this balance lowers performance due to increased reasoning demands. Qwen2VL follows a similar pattern but achieves higher accuracy, likely due to visual training.
GPT-o4-mini with high visual reasoning shows the same pattern, which is later discussed as an additional experiment in Appendix~\ref{appendix:gpt-o4-mini-high}.
Moreover, our experiments with humans, presumed to have prior visual knowledge, show strong reasoning ability, achieving over 90\% accuracy even in scenarios where perspective preference requires FoR adaptation.
In our results, humans slightly prefer intrinsic FoR over relative FoR. Their cultural background and recently seen examples may also influence their preference as discussed in~\cite{cueBySeenAndHear}. A detailed analysis of human performance is provided in Appendix~\ref{appendix:human_study}.
In summary, we conjecture that Qwen2 performs well by focusing on extraction without reasoning FoR, while other models attempt reasoning but often fail to reach correct conclusions, leading to lower accuracy. To support this claim and further analyze the results, we provide a quantitative analysis in Appendix~\ref{appendix:quantitative_analysis}, examining how linguistic expressions of spatial relations and facing directions in the context influence model outputs.

\begin{figure}[t]
    \centering
    \includegraphics[width=0.9\linewidth]{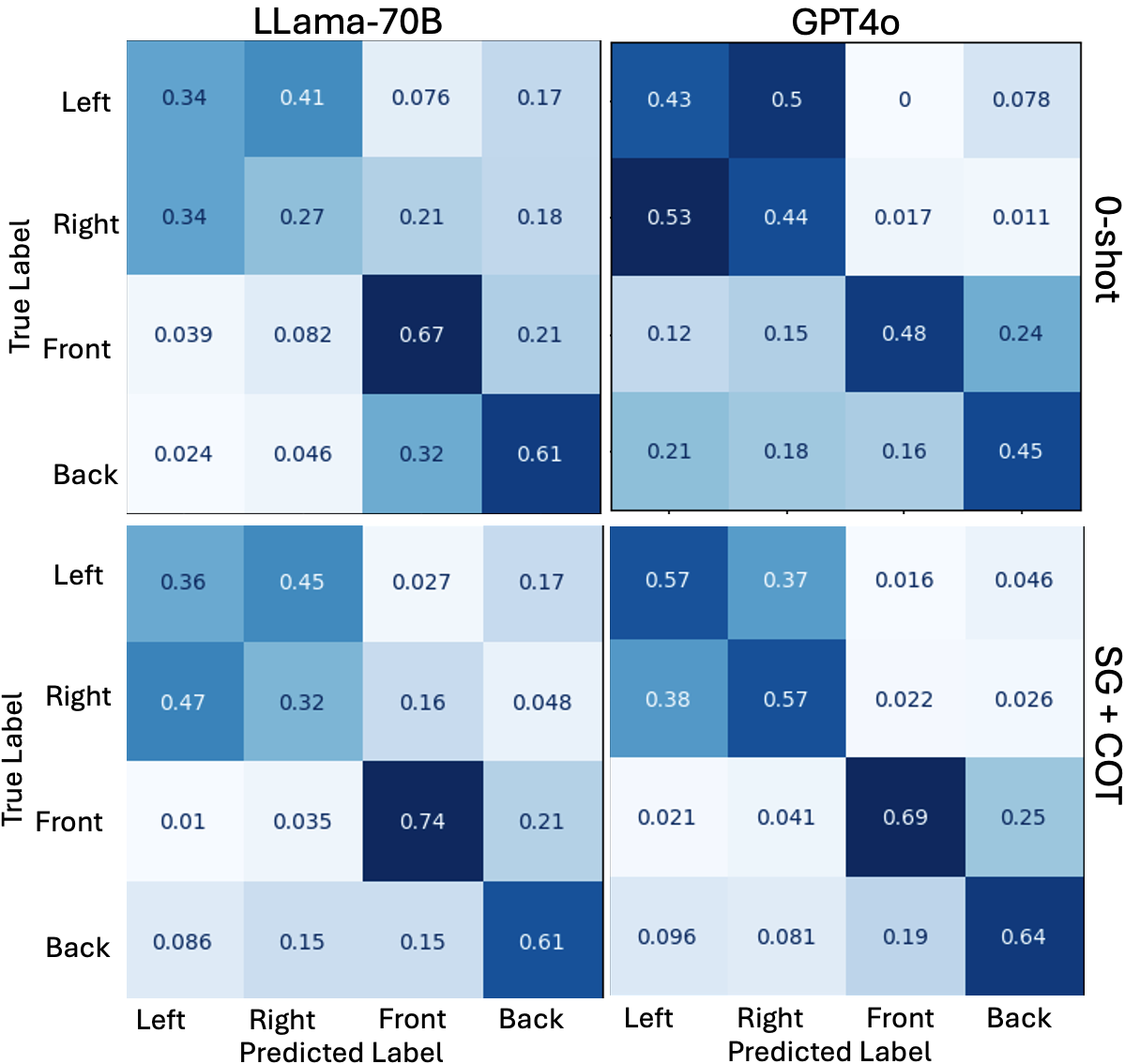}
    \caption{Confusion matrices of spatial relation predictions by Llama3 and GPT-4o in 0-shot and SG+CoT settings, when FoR adaptation is required.}
    \label{fig:cf_conversion}
\end{figure}

\noindent\textbf{RQ2. Can LLMs adapt FoR when answering the questions?}
To address this question, we analyze QA results of C-split in Table~\ref{tab:QA_c_split}, where context and question explicitly specify the perspective.
Results show that LLMs struggle with FoR conversion, especially when the question is asked from the relatum’s perspective and the context is given from the camera’s perspective, with the highest accuracy only 55.5\% with Llama3-70B (CoT).
We further analyze FoR adaptation in Llama3-70B and GPT-4o using the confusion matrix in Figure~\ref{fig:cf_conversion}.
Our findings reveal that pure-text LLM, Llama3-70B, systematically reverses left and right. 
This contrasts with humans, who in English typically reverse front and back when describing spatial relations from a perspective, while preserving lateral directions~\cite{Hill1982}.
This difference explains Llama3’s poor adaptation to the camera perspective.
In contrast, large multimodal models like GPT-4o follow expected patterns, consistent with~\citealt{comfortFoR}.
While the GPT-4o results suggest some ability to convert the relatum’s perspective into the camera’s with in-context learning (up to 82.4\% accuracy), the reverse direction—from the camera’s perspective to the relatum’s—remains challenging, reaching only 53\% accuracy (see GPT-4o+CoT). 
A similar trend appears when comparing Qwen2 and Qwen2VL, as discussed in Appendix~\ref{appendix:QwenDiscussion}. 
Qwen2VL performs better when shifting to the camera perspective but performs worse with the reverse direction, often failing to generate correct reasoning. This difficulty persists for several models when converting spatial relations from images to the relatum's perspective, as noted in~\citealt{comfortFoR}.
Based on our Attention analysis, one possible reason for poor performance in perspective shift in Qwen2 is that the model often overlooks orientation tokens, critical for answering FoR questions. The detail of the analysis and visualizations can be found in Appendix~\ref{appendix:attention}. 
Another observed reason is that the model tends to rely on a fixed facing direction when answering (facing toward or facing away). We provide a quantitative analysis of this behavior in Appendix~\ref{appendix:quantitative_analysis}.

\noindent\textbf{RQ3. How can an explicit FoR identification help spatial reasoning in QA?}
We compare CoT and SG+CoT results to assess the effect of FoR identification on LLMs’ spatial reasoning in QA.
Based on C-Split results (Table~\ref{tab:QA_c_split}), incorporating SG improves the model’s ability to identify the correct perspective from input expression ranging from 2.9\% to 30\% of cases where the context and question share the same perspective. 
These cases are easier as the models do not need FoR adaptation. 
However, two notable exceptions arise.
First, Llama3 performs poorly on camera-perspective questions, and FoR identification via SG fails to improve its performance. This may be due to Llama3’s lack of visual training, which we speculate limits its FoR understanding.
Second, Qwen2VL struggles with relatum-perspective reasoning, showing negative gains even with CoT.
SG is less effective when context and question differ in perspective; while it helps identify the correct FoR in context, it does not enhance reasoning across perspectives.
This limitation is evident in A-Split results (Table~\ref{tab:A_split-QA}), where models only improve significantly when SG aligns their preference with the question’s perspective, as seen in Qwen2-72B and GPT-4o.
SG identification results are reported in the Appendix~\ref{appendix:FoRIdentification}.
Still, FoR identification improves overall spatial reasoning (see Avg. column for SG+CoT in Table~\ref{tab:I_split}).

\noindent\textbf{RQ4. How can explicit FoR identification help spatial reasoning in visualization?}
To address this question, we evaluate the SG+GLIGEN baseline with a focus on the VISOR$_{cond}$ metric, which better reflects the model’s spatial understanding than the overall performance measured by VISOR$_{uncond}$ alternative that is reported in Appendix~\ref{appendix:Visor_uncond}. 
Table~\ref{tab:I_split} shows that adding spatial information and FoR classes (SG+GLIGEN) improves performance across all splits compared to the GLIGEN baseline.
In particular, SG improved the model's performance when expressions follow a relative FoR. This finding aligns with QA results in Table~\ref{tab:A_split-QA}, where \textit{Llama3 prefers relative FoR in camera-perspective scenarios}.
In contrast, baseline diffusion models (SD-1.5 and SD-2.1) perform better for intrinsic FoR, even though GLIGEN is based on SD-2.1.
This outcome is likely due to GLIGEN's reliance on bounding boxes for spatial configurations, which limits handling of intrinsic FoR because object properties and orientation—critical for intrinsic reasoning—are missing. Despite this limitation, incorporating FoR information via SG-prompting improves performance across all FoR classes.
We provide further analysis on SG for the layout generation in Appendix~\ref{appedix:anaylize_SG_improment_t2i}.

\noindent\textbf{RQ5. How does human FoR understanding compare to LLMs'?}
For understanding FoR, humans usually adopt the perspective of a specific object, a process linked to the theory of mind, when describing spatial relations~\cite{LoyDemberg2023}. However, the ability to shift spatial perspective can be influenced by factors such as social context~\cite{RegehrGagnonGeussStefanucci2013}, situational circumstances~\cite{GunalpMoossaianHegarty2019}, and linguistic or cultural background~\cite{BohnemeyerEtAl2014}. These factors highlight the inherent difficulty of FoR reasoning, even for humans.
For LLMs, the challenge is even more pronounced. A key limitation stems from the nature of their training data, which is often based on image–caption pairs from 2D images or spatial descriptions grounded primarily in the visual modality from a camera perspective.
As a result, models tend to learn spatial relations only from this specific viewpoint, restricting their ability to generalize beyond it. 
As shown in Table~\ref{tab:A_split-QA}, our results indicate that most language-only models (e.g., Llama3) struggle with perspective shifts, whereas multimodal LLMs (e.g., GPT-4o) perform significantly better, though primarily from the camera perspective.
This bias of MLLMs toward a single FoR system is consistent with previous findings~\cite{comfortFoR}, which show that MLLMs often fail to adapt to cultural variation in perspective-taking, typically aligning with the English language.
This bias underscores the need to improve LLMs’ ability to generalize across diverse spatial reasoning tasks.
Future work would be interesting to explore novel ideas for both large-scale and realistic training data, as well as reasoning techniques that enable models' FoR reasoning.




\section{Related Work}
\noindent\textbf{Frame of Reference in Cognitive-Linguistic (CL) Studies.}
The concept of the frame of reference in CL studies was introduced by \citealt{Levinson_2003} and later expanded with more diverse spatial relations \citep{TENBRINK2011704}.
Subsequent research investigated the human preferences for specific FoR classes~\citep{VUKOVIC2015110, SHUSTERMAN2016115, Ruotolo2016, cueBySeenAndHear}. For instance, \citealt{Ruotolo2016} examined how FoR influences scene memorization. They found that participants performed better when spatial relations were based on their own position rather than external objects, highlighting a distinction between relative and intrinsic FoR. 

\noindent\textbf{Frame of Reference in AI.}
Several benchmarks have been developed to evaluate the spatial understanding of AI models in multiple modalities; for instance, 
textual QA~\citep{shi2022stepgamenewbenchmarkrobust, mirzaee2022transferlearningsyntheticcorpora, rizvi2024sparcsparpspatialreasoning}, and text-to-image (T2I) benchmarks~\citep{gokhale2023benchmarkingspatialrelationshipstexttoimage, cho2023dallevalprobingreasoningskills, cho2023visualprogrammingtexttoimagegeneration}.
However, most of these benchmarks overlook FoR, assuming a single FoR for all instances despite its significance in cognition.
Recent vision-language studies have begun addressing this gap~\cite{liu2023visualspatialreasoning, comfortFoR, zhang2025spartund, wang2025spatial457}. 
For instance,\citealt{zhang2025spartund, wang2025spatial457} propose benchmarks that incorporate the concept of perspective in embodied AI and situated 3D environments, although FoR is not the primary focus of either work.
\citealt{liu2023visualspatialreasoning} examines FoR’s impact on visual question answering but focuses only on the intrinsic and relative FoR categories. Our work covers a wider range of FoRs.
\citealt{comfortFoR} explores FoR ambiguity by evaluating spatial relations from camera-perspective images, with FoR specified in the question. In contrast, our work examines spatial reasoning across multiple FoRs and perspective changes, extending beyond the camera’s viewpoint. We further demonstrate that explicitly identifying FoR for in-context learning enhances spatial reasoning in both QA and T2I tasks.

\section{Conclusion}
Given the significance of spatial reasoning in AI models, we introduce the \textbf{F}rame \textbf{o}f \textbf{R}eference \textbf{E}valuation in \textbf{S}patial Reasoning \textbf{T}asks (FoREST) benchmark to evaluate FoR comprehension in textual spatial expressions through question answering and grounding in the visual modality via diffusion models for text to image generation. Using this benchmark, we identify notable differences in FoR comprehension across LLMs, as well as their struggles with questions that require adapting between multiple FoRs. Moreover, biases in FoR interpretation affect layout generation in text-to-image models. To improve FoR comprehension, we propose Spatial-Guided prompting, which extracts topological, distal, and directional information in addition to FoR, and incorporates this knowledge into downstream prompting. Employing SG improves performance in both QA tasks requiring FoR understanding and text-to-image generation models by providing a more accurate layout to these models.

\section*{Acknowledgment}
This project is partially supported by the Office of Naval Research (ONR) grant N00014-23-1-2417. Any opinions, findings, and conclusions or recommendations expressed in this material are those of the authors and do not necessarily reflect the views of Office of Naval Research. We thank anonymous reviewers for their constructive feedback, which greatly helped us improve this manuscript.

\section*{Limitations}
While we analyze LLMs' shortcomings, our benchmark only highlights areas for improvement.
The trustworthiness and reliability of the LLMs are still a research challenge.
Our analysis is confined to the spatial reasoning domain and does not account for biases related to gender or race. We acknowledge that linguistic and cultural variations in spatial expression are not considered, as our study focuses solely on English. Extending this work to multiple languages could reveal important differences in FoR adaptation. 
Our analysis is still limited to the synthetic environment. Future research should consider the broader implications of the frame of reference of spatial reasoning in real-world applications.
Additionally, our experiments require substantial GPU resources, limiting the selection of LLMs and constraining the feasibility of testing larger models. The computational demands also pose accessibility challenges for researchers with limited resources.
We find no ethical concerns in our methodology or results, as our study does not involve human subjects or sensitive data.

\bibliography{ref}

\appendix
\section{Additional Details of FoREST Creation}\label{appendix:dataset_creation}
\begin{figure*}[t]
    \centering
    \includegraphics[width=0.8\linewidth]{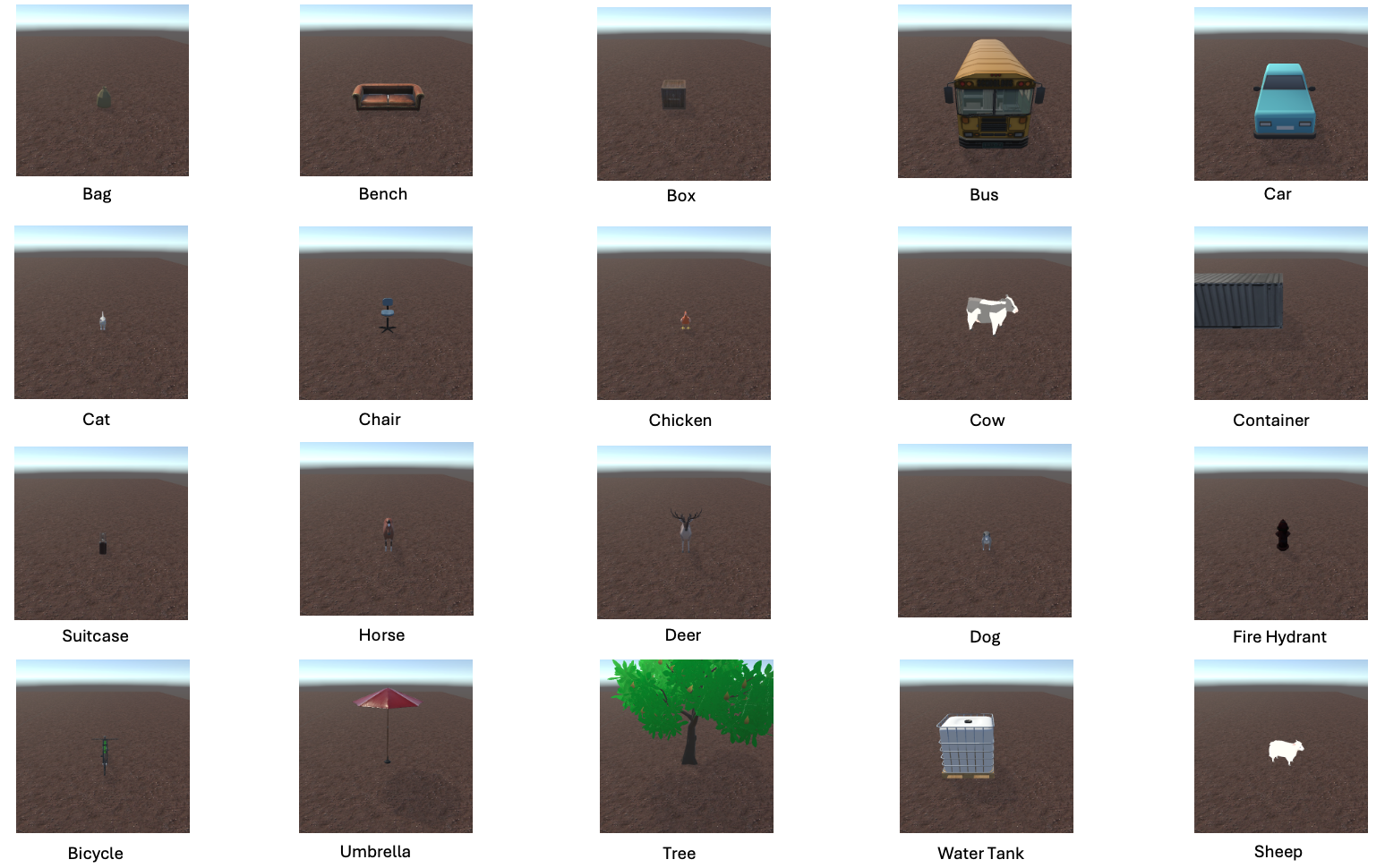}
    \caption{All 3d models used to generate visualizations for FoREST.}
    \label{fig:3D_model}
\end{figure*}

We define the nine categories of objects selected in our dataset as indicated below in Table~\ref{tab:selected_object}. We select sets of locatum and relatum based on the properties of each class to cover four cases of frame of reference defined in Section~\ref{sec:FoR_Relatum_scenario}. Notice that we also consider the appropriateness of the container; for example, the car should not contain the bus.
Based on the selected locatum and relatum. To create an A-split spatial expression, we substitute the actual locatum and relatum objects in the Spatial Relation template. After obtaining the A-split contexts, we create their counterparts using the perspective/topology clauses to make the counterparts in the C-split. Then, we obtain the I-A and I-C split by applying the directional template to the first occurrence of relatum when it has intrinsic directions. The directional templates are ``that is facing towards,'' ``that is facing backward,'' ``that is facing to the left,'' and "that is facing to the right.'' All the templates are in the Table~\ref{tab:templates}.
We then construct the scene configuration from each modified spatial expression and send it to the simulator developed using Unity3D. 
Eventually, the simulator produces four visualization images for each scene configuration. 

\begin{table*}[t]
    \centering
    \small
    \begin{tabular}{ c c c c }
    \toprule
        Category &  Object(s) & Intrinsic Direction & Possible Container \\
        \hline
        Small object without intrinsic directions & umbrella, bag, suitcase, fire hydrant & \xmark & \xmark \\
        \hline
        Big object with intrinsic directions & bench, chair & \checkmark &  \xmark \\ 
        \hline
        Big object without intrinsic direction & water tank & \xmark & \xmark \\
        \hline
        Container & box, container & \xmark & \checkmark \\
        \hline
        Small animal & chicken, dog, cat & \checkmark & \xmark \\
        \hline
        Big animal & deer, horse, cow, sheep & \checkmark & \xmark \\
        \hline
        Small vehicle & bicycle & \checkmark & \xmark \\
        \hline
        Big vehicle & bus, car & \checkmark & \checkmark \\
        \hline
        Tree & tree & \xmark & \xmark \\
        \bottomrule
    \end{tabular}
    \caption{All selected objects with two properties: intrinsic direction, and affordance of being a container}
    \label{tab:selected_object}
\end{table*}

\subsection{Scene generation}
The process begins by randomly placing the relatum in the scene with an orientation specified by the scene configuration. The relatum’s orientation is then sampled from predefined ranges: [-40, 40] for front, [40, 140] for left, [140, 220] for back, and [220, 320] for right. Next, the locatum is positioned relative to the relatum according to the given spatial relation. If the FoR is relative, the locatum is placed with respect to the camera’s orientation; otherwise, it is placed with respect to the relatum’s orientation. We then check whether both objects are visible from the camera. If not, the process regenerates the locatum and the relatum until a valid placement is achieved. Once placement is finalized, one of six backgrounds is randomly selected. This procedure is repeated four times for each scene configuration.

\subsection{Object models and background}
For the object models and background, we obtain them from the Unity Asset Store\footnote {https://assetstore.unity.com}. All of them are free and available for download. All 3D models used are shown in Figure~\ref{fig:3D_model}.

\begin{table*}[t]
    \centering
    \small
    \begin{tabular}{ c | l } 
    \toprule
        &   \{locatum\} is in front of \{relatum\} \\
         &  \{locatum\} is on the left of \{relatum\} \\
           &  \{locatum\} is to the left of \{relatum\}\\ 
        Spatial Relation Templates &  \{locatum\} is behind of \{relatum\} \\
           &  \{locatum\} is back of \{relatum\} \\
          &  \{locatum\} is on the right of \{relatum\} \\
          &   \{locatum\} is to the right of \{relatum\} \\
         \hline
         &  within \{relatum\} \\ 
         Topology Templates &  and inside \{relatum\} \\ 
         &  and outside of \{relatum\} \\ 
         \hline
         & from \{relatum\}'s view \\ 
         & relative to \{relatum\} \\ 
         Perspective Templates & from \{relatum\}'s perspective\\ 
         & from my perspective \\ 
          & from my point of view \\ 
           & relative to observer \\ 
         \hline
          &  \{relatum\} facing toward that camera\\ 
         Orientation Templates &  \{relatum\}is facing away from the camera. \\ 
          &  \{relatum\} facing left relative to the camera\\ 
           &  \{relatum\} facing right relative to the camera \\ 
         \bottomrule
    \end{tabular}
    \caption{All templates used to create FoREST dataset.}
    \label{tab:templates}
\end{table*}

\subsection{Templates}
\label{appendix:textual_template}

{\textbf{Context templates}} All manually created templates used to create a FoREST spatial expression are given in Table~\ref{tab:templates}.

\begin{table*}[t]
    \centering
    \small
    \setlength{\tabcolsep}{0.9mm}
    \begin{tabular}{c|l} 
    \toprule
         & In the \{perspective\}, how is \{locatum\} positioned in relation to \{relatum\}? \\
        &  Based on the \{perspective\}, where is the \{locatum\} from the \{relatum\}'s position? \\ 
         Normal Templates &  From the \{perspective\}, what is the relation of the\{locatum\} to the \{relatum\}?\\ 
          &  Looking through the \{perspective\}, how does \{locatum\} appear to be oriented relative to \{relatum\}'s position?\\ 
           &  Based on the \{perspective\}, where is \{locatum\} located with respect to \{relatum\}'s location? \\ 
           \midrule
                    & In relation to the \{relatum\}, where is the \{locatum\} located when viewed from \{perspective\}? \\
        &  In the \{perspective\}, given \{relatum\} as reference, where is \{locatum\} located? \\ 
         Reverse Templates &  Relative to the \{relatum\}, where can the \{locatum\} be found from the viewpoint of \{perspective\}?\\ 
          &  Considering \{relatum\} as reference, where does \{locatum\} lie when seen from \{perspective\}?\\ 
         \bottomrule
    \end{tabular}
    \caption{Question templates used to construct the FoREST dataset. Normal templates refer to cases where the locatum precedes the relatum, while reversed templates are the opposite. Normal templates are used in the main experiments, and reversed templates in the additional experiment (Appendix~\ref{appendix:inverted_question}).}
    \label{tab:GPT_templates}
\end{table*}

\noindent\textbf{Question Templates.} To generate question templates, we first prompt GPT-4o with a manually created template (the first in Table~\ref{tab:GPT_templates}). GPT-4o then generates eight additional variations, including versions with the relatum and locatum in reversed order. Each template was manually reviewed and validated before being added to our corpus; the full set is listed in Table~\ref{tab:GPT_templates}. Templates from the camera’s perspective use the camera as the {perspective}, while templates from the relatum’s perspective use the relatum object.
For each generated scenario, we select one normal and one reversed template to ensure both orders are represented, allowing us to evaluate whether the order of the locatum and relatum influences model performance. In the FoREST dataset, the contexts of both orders are kept separate.

\section{Impact of Visual Training}\label{appendix:QwenDiscussion}

\begin{figure}
    \centering
    \includegraphics[width=\linewidth]{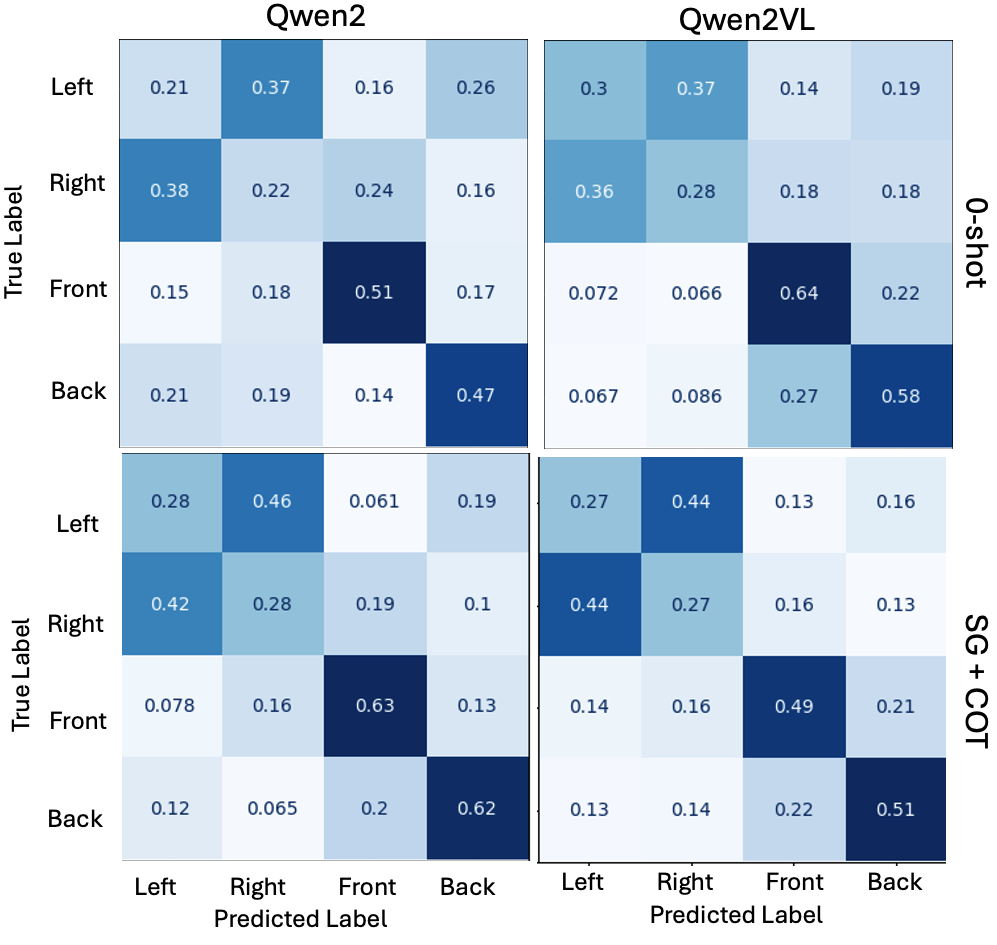}
    \caption{Confusion matrices of spatial relation answers when Qwen2 and Qwen2-VL must adapt FoR in the 0-shot and (SG+CoT) settings.}
    \label{fig:enter-label}
\end{figure}

To analyze the impact of visual training, we compare Qwen2 with Qwen2-VL, which extends Qwen2 with visual perception capability. In the A-split (Table~\ref{tab:A_split-QA}), Qwen2-VL shows a more balanced preference pattern and stronger performance than Qwen2. This suggests that, unlike Qwen2, which often extracts spatial relations without accounting for perspective, Qwen2-VL can also reason about perspective shifts.
This observation is further supported by the C-split results in Table~\ref{tab:QA_c_split}, where Qwen2-VL substantially outperforms Qwen2 in cases requiring adaptation of spatial relations across perspectives. These findings reinforce our hypothesis that visual training improves a model’s ability to interpret perspective changes.
Nevertheless, Qwen2-VL still struggles to reason from the relatum’s perspective in CoT and SG+CoT settings. While visual training enables it to handle perspective shifts more effectively, its reasoning remains more accurate when questions are framed from the camera’s viewpoint. This limitation is expected, since training data—particularly image-captioning datasets—are typically annotated from a human, camera-based perspective, restricting the model’s ability to generalize across viewpoints.

\section{GPT-o4-mini-high Results}\label{appendix:gpt-o4-mini-high}
For comparison with the main paper results, we include newer modelss such as GPT-4o-mini-high, a model configured for enhanced reasoning. As shown in Table~\ref{tab:A_split-QA-o4-mini}, GPT-4o-mini-high exhibits a balanced preference between relative and intrinsic interpretations in the A-split. Similar to Qwen2VL, it demonstrates strong reasoning ability, achieving high accuracy even when FoR adaptation is required, likely due to its visual reasoning capability. Notably, in the C-split, GPT-4o-mini-high performs well on relatum-perspective questions, even outperforming GPT-4o (SG+CoT) in Table~\ref{tab:QA_c_split-o4-mini}. However, its performance declines on camera-perspective questions, perhaps due to training that emphasizes alternative perspectives over the camera view.  These results indicate that perspective shifting remains a challenge in the textual domain, even for strong reasoning models.

\begin{table*}[t]
    \setlength{\tabcolsep}{1mm}
    \small
    \centering
    \begin{tabular}{ l  c c  c | c c  c | c | c | c  | c c  c | c c  c| c }
    \toprule
     & \multicolumn{9}{c|}{Question with Camera Perspective} & \multicolumn{7}{c}{Question with Relatum perspective} \\
    \cline{2-17}
     Model & \multicolumn{3}{c|}{Cow} & \multicolumn{3}{c|}{Car} & Box & Pen & Avg. &\multicolumn{3}{c|}{Cow} & \multicolumn{3}{c|}{Car} & Avg. \\
      \cline{2-17}
       & R\% & I\% & Acc. & R\% & I\% & Acc. & Acc. & Acc. & Acc. &R\% & I\% & Acc. & R\% & I\% & Acc. & Acc. \\ 
\hline
GPT-4o (1) & $\mathbf{84.3}$ & $15.3$ & $94.5$ & $\mathbf{88.5}$ & $11.0$ & $97.3$ & $99.2$ & $99.8$ & $95.6$  & $21.6$ & $\mathbf{78.0}$ & $91.6$ & $16.1$ & $\mathbf{83.5}$ & $90.5$ & $91.4$ \\
GPT-4o (2) & $\mathbf{69.0}$ & $30.6$ & $76.6$ & $\mathbf{80.3}$ & $19.2$ & $89.5$ & $100.0$ & $100.0$ & $81.5$  & $29.0$ & $\mathbf{70.5}$ & $74.7$ & $30.9$ & $\mathbf{68.7}$ & $77.5$ & $75.1$ \\
GPT-4o (3) & $41.5$ & $\mathbf{58.3}$ & $92.3$ & $38.2$ & $\mathbf{61.6}$ & $91.0$ & $100.0$ & $99.8$ & $93.2$  & $33.9$ & $\mathbf{65.8}$ & $93.9$ & $32.0$ & $\mathbf{67.6}$ & $93.9$ & $93.9$ \\
GPT-4o (4) & $26.0$ & $\mathbf{73.9}$ & $79.2$ & $27.7$ & $\mathbf{72.1}$ & $79.4$ & $96.7$ & $94.3$ & $81.4$  & $16.2$ & $\mathbf{83.4}$ & $95.5$ & $19.2$ & $\mathbf{80.4}$ & $94.8$ & $95.4$ \\
\hline
o4-mini-high & $\mathbf{68.0}$ & $31.6$ & $92.0$ & $\mathbf{69.9}$ & $29.7$ & $94.7$ & $100.0$ & $99.4$ & $93.4$  & $\mathbf{58.5}$ & $41.3$ & $86.9$ & $\mathbf{50.2}$ & $49.5$ & $93.3$ & $87.8$ \\
\hline
Human-baseline & $36.6$ & $\mathbf{63.4}$ & $90.7$ & $27.8$ & $\mathbf{72.2}$ & $96.0$ & $72.0$ & $82.7$ & $85.3$ & $41.4$ & $\mathbf{58.6}$ & $97.3$ & $36.1$ & $\mathbf{63.9}$ & $96.0$ & $96.7$ \\
\bottomrule
    \end{tabular}
    \caption{Additional results of QA accuracy in the A-Split with GPT-o4-mini-high. R\% and I\% represent the percentage the model assumes relative or intrinsic FoR for ambiguous expression, explained in Section~\ref{sec:evaluation_setting}. Acc is the accuracy, and Avg is the micro-average of accuracy. (1): 0-shot, (2): 4-shot, (3): CoT, and (4): SG+CoT.}
    \label{tab:A_split-QA-o4-mini}
\end{table*}

\begin{table*}[ht!]
    \small
    \setlength{\tabcolsep}{1mm}
    \centering
    \begin{tabular}{ l c c c c c |c c c c c }
    \toprule
     & \multicolumn{5}{c|}{Question with Camera Perspective} & \multicolumn{5}{c}{Question with Relatum Perspective} \\
    \cline{2 - 11}
    Model & ER (CP) & EI (RP) & II (RP) & IR (CP) & Avg. & ER (CP) & EI (RP) & II (RP) & IR (CP) & Avg. \\
    \hline
    GPT-4o (0-shot)  & $79.7$ & $45.1$ & $39.5$ & $90.2$ & $64.2$  & $46.9$ & $88.5$ & $98.2$ & $34.8$ & $67.5$ \\
    GPT-4o (4-shot) & $68.0$ & $52.6$ & $60.7$ & $74.1$ & $61.8$  & $44.9$ & $\mathbf{98.2}$ & $\mathbf{100.0}$ & $37.5$ & $71.2$ \\
    GPT-4o (CoT) & $81.7$ & $\mathbf{76.1}$ & $\mathbf{82.4}$ & $71.5$ & $78.8$  & $53.0$ & $91.1$ & $90.6$ & $50.8$ & $71.9$ \\
    GPT-4o (SG+CoT)  & $\mathbf{97.9}$ & $72.2$ & $72.7$ & $\mathbf{93.4}$ & $\mathbf{85.8}$  & $48.9$ & $96.3$ & $95.9$ & $36.1$ & $71.8$ \\
    \hline
    o4-mini-high & $68.7$ & $73.1$ & $72.1$ & $79.0$ & $71.4$  & $\mathbf{78.9}$ & $91.9$ & $93.9$ & $\mathbf{62.7}$ & $\mathbf{84.5}$ \\
\bottomrule
    \end{tabular}
    \caption{Additional results of QA accuracy in the C-Split with GPT-o4-mini-high. ER, EI, II, and IR  denote external relative, external intrinsic, internal intrinsic, and internal relative FoRs. Avg represents the micro-average accuracy. CP refers to context with camera perspective, while RP denotes context with relatum perspective.}
    \label{tab:QA_c_split-o4-mini}
\end{table*}

\section{Attention Analysis}\label{appendix:attention}
\begin{figure}[t]
    \centering
    \includegraphics[width=0.9\linewidth]{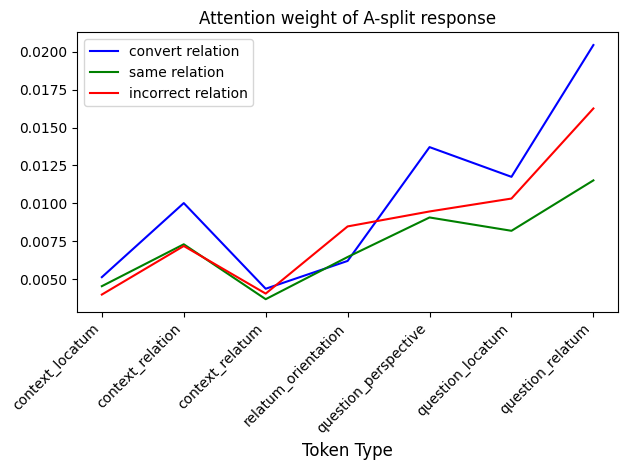}
    \caption{Attention weights of input context for Question Answering in A-split. Considering the following semantic concepts: locatum mention in context, relation in context, relatum in context, relatum orientation, perspective in question, locatum in question, and relatum in question. Three patterns are considered: (1) incorrect answers, (2) correct answers assuming a shared perspective between context and question, and (3) correct answers assuming different perspectives.}
    \label{fig:attent_weight_A}
\end{figure}

\subsection{Experimental setting}
We conducted an attention-based interpretability analysis using Qwen2 in a 0-shot setting, where only the context and question were provided.
We focused on analyzing attention weights across key semantic elements in both A-split and C-split.
We analyzed the attention weights for various groups of tokens that convey the following concepts: locatum mention occurring in the context, relation in context, relatum in context, perspective in context, relatum orientation, perspective in question, locatum in question, and relatum occurring in the question.
We categorized response patterns into three scenarios: (1) the model answers incorrectly, (2) the model answers correctly with the assumption that the same perspective holds for context and question, and (3) the model answers correctly with the assumption that context and question have different perspectives. Note that we separate the analysis for both splits because the conceptual tokens are different, that is, the perspective mentioned in the context is only for the C-split.

\subsection{Experimental results}

\noindent\textbf{A-split.} 
According to Figure~\ref{fig:attent_weight_A}, we found that, for the correct responses, the model paid the least attention to the question's perspective in the \textit{same perspective}. Meanwhile, for the \textit{different perspective} case, attention significantly increased to tokens related to the relation, the question perspective, and the relatum in the question. 
This result illustrates that these tokens played an important role in adapting spatial relations for perspective shifts. 
However, when the model increased attention weight on relatum orientation compared to other tokens, more errors occurred in answering the questions.

\begin{figure*}[t]
    \centering
    \begin{subfigure}[ht]{0.45\textwidth}
        \centering
        \includegraphics[width=\textwidth, trim={0 0 0 2cm}]{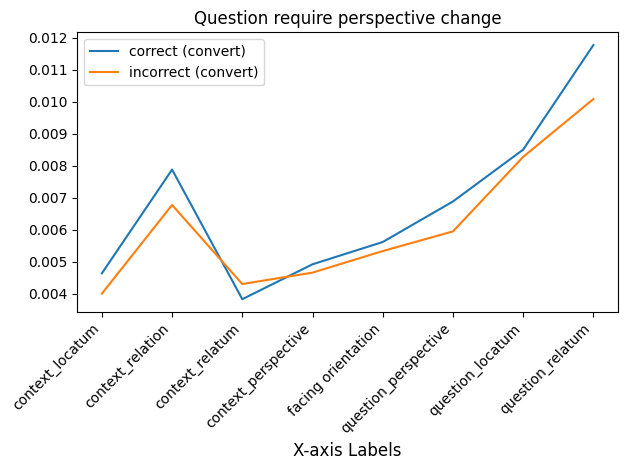}
        \caption{Questions that require changing perspective.}
    \end{subfigure}%
    ~ 
    \begin{subfigure}[ht]{0.45\textwidth}
        \centering
        \includegraphics[width=\textwidth, trim={0 0 0 2cm}]{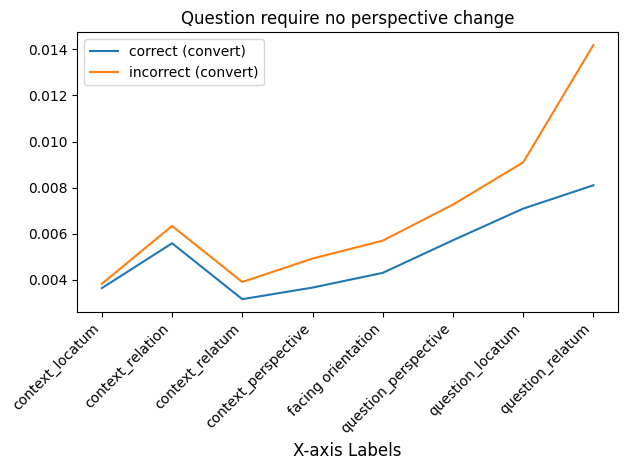}
        \caption{Questions that do not require changing perspective.}
    \end{subfigure}

    \caption{Attention weights of input context for Question Answering in C-split. Considering the following semantic concepts: locatum mention in context, relation in context, relatum in context, perspective in context, relatum orientation, perspective in question, locatum in question, and relatum in question. Two patterns are considered: (1) incorrect answers, (2) correct answers. We separate the questions that require and do not require perspective changing.}
    \label{fig:attent_weight_C}
\end{figure*}

\noindent\textbf{C-split.}
According to Figure~\ref{fig:attent_weight_C}, in cases that require a shift in perspective, we observed that the model pays significantly more attention to the \textit{spatial relation in context}, \textit{the question perspective}, and \textit{the relatum in the question} concepts necessary for answering correctly.
However, the model largely overlooks the orientation tokens, which also play a crucial role in resolving these questions. 
This lack of attention to orientation may majorly contribute to the model’s failures in perspective-shifting scenarios. 
In cases where the relation mentioned remains the same in both the context and the answer, we observed that the model, in incorrect predictions, tends to assign higher attention to the relatum in the question. 
In contrast, correct predictions are associated with more balanced attention across tokens. Maintaining moderate attention to all tokens appears to help the model consistently respond with the correct relation, repeating it from the context.

\section{ Analysis of Various Spatial Configurations }\label{appendix:quantitative_analysis}
In this section, we performed additional diagnostics by analyzing key features that may contribute to model failures, including the facing direction of the relatum of C-split and the type of spatial relations expressed in the context. This analysis is conducted over two settings: 0-shot and SG+CoT, using both Qwen2-72B and GPT-4o models.

\subsection{Facing direction}

\begin{table*}[ht]
    \centering
    \setlength{\tabcolsep}{1.2mm}
    \small
    \begin{tabular}{l c c c c c | l c c c c c c}
    \toprule
    \multicolumn{6}{ c |}{Qwen2-72B + 0-shot} & \multicolumn{6}{ c }{GPT-4o + 0-shot}  \\
    \hline
    Facing Direction & Front & Back & Left & Right & Avg. & Facing Direction & Front & Back & Left & Right & Avg. \\
    \hline
    Front & 83.97 & 80.53 & 33.21 & 33.78 & 57.87 & Front & 84.54 & 77.29 & 66.79 & 69.85 & 74.62 \\ 
    Back & 70.42 & 66.79 & 29.20 & 32.82 & 49.81 & Back & 58.02 & 47.52 & 16.41 & 30.15 & 38.02 \\ 
    Left & 0.19 & 0.00 & 2.29 & 39.50 & 10.50 & Left & 96.95 & 61.64 & 5.73 & 9.92 & 43.56 \\ 
    Right & 0.00 & 0.00 & 26.15 & 3.63 & 7.44 & Right & 64.31 & 26.34 & 1.53 & 1.72 & 23.47 \\
    \midrule
    \multicolumn{6}{ c |}{Qwen2-72B + SG + CoT} & \multicolumn{6}{ c }{GPT-4o +  SG + CoT}  \\
    \hline
    Facing Direction & Front & Back & Left & Right & Avg. & Facing Direction & Front & Back & Left & Right & Avg. \\
    \hline
    Front & 98.09 & 83.02 & 6.87 & 13.17 & 50.29 & Front & 50.95 & 58.78 & 95.80 & 95.99 & 75.38 \\ 
    Back & 75.00 & 75.57 & 37.60 & 36.83 & 56.25 & Back & 86.26 & 93.89 & 2.86  & 2.29  & 46.33 \\ 
    Left &  12.21 & 19.27 & 0.38  & 83.78 & 28.91 & Left & 73.47 & 35.11 & 51.34 & 63.17 & 55.77 \\ 
    Right & 5.73  & 48.28 & 57.25 & 9.73  & 30.25 & Right & 79.01 & 56.49 & 60.31 & 49.81 & 61.40 \\
    \bottomrule
    \end{tabular}
    \caption{The accuracy of each relation type (column-label) for each facing direction (row-label) presented in the input context for Qwen2-72B and GPT-40 with 0-shot and SG+CoT settings.}
    \label{tab:facing_direction_diagnostic}
\end{table*}

According to Table~\ref{tab:facing_direction_diagnostic}, we observe that the facing direction plays a significant role in the performance drop for both Qwen2-72B and GPT-4o. Both models achieve relatively high performance only when the relatum is facing either toward or away from the camera, compared to when it is facing left or right relative to the camera. This effect is particularly noticeable in GPT-4o, which shows a substantial performance gap between cases where the relatum is facing the camera and other directions. We speculate that this error stems from biases in the training data, which typically consist of image–caption pairs based on 2D images where objects are clearly oriented toward the camera. As a result, LLMs appear to overfit to FoR reasoning patterns on specific facing directions (facing toward the camera), leading to confusion—particularly between “left” and “right” that usually reverse when the relatum is facing toward the camera—when the relatum is oriented differently, even if the image is not presented in the input.

\subsection{Spatial relation}
\begin{table*}[ht]
    \centering
    \setlength{\tabcolsep}{1.2mm}
    \small
    \begin{tabular}{l c c c c c | l c c c c c c}
    \toprule
    \multicolumn{6}{ c |}{Qwen2-72B + 0-shot} & \multicolumn{6}{ c }{GPT-4o + 0-shot}  \\
    \hline
    Spatial Relation  & Front & Back & Left & Right & Avg. & Facing Direction & Front & Back & Left & Right & Avg. \\
    \hline
    Front & 83.97 & 66.79 & 2.29  & 3.63  & 39.17 & Front & 84.54 & 47.52 & 5.73  & 1.72  & 34.88 \\ 
    Back & 70.42 & 80.53 & 26.15 & 39.50 & 54.15 & Back & 58.02 & 77.29 & 1.53 & 9.92 & 36.69 \\ 
    Left & 0.19  & 0.00  & 29.20 & 33.78 & 15.79 & Left & 96.95 & 26.34 & 16.41 & 69.85 & 52.39  \\ 
    Right & 0.00  & 0.00  & 33.21 & 32.82 & 16.51 & Right & 64.34 & 61.64 & 66.79 & 30.15 & 55.73 \\
    \midrule
    \multicolumn{6}{ c |}{Qwen2-72B + SG + CoT} & \multicolumn{6}{ c }{GPT-4o +  SG + CoT}  \\
    \hline
    Spatial Relation & Front & Back & Left & Right & Avg. & Facing Direction & Front & Back & Left & Right & Avg. \\
    \hline
    Front & 98.09 & 75.57 & 0.38  & 9.73  & 45.94 & Front & 50.95 & 93.89 & 51.34 & 49.81 & 61.50 \\ 
    Back & 75.00 & 83.02 & 57.25 & 83.78 & 74.76 & Back & 86.26 & 58.78 & 60.31 & 63.17 & 67.13 \\ 
    Left & 12.21 & 48.28 & 37.60 & 13.17 & 27.81 & Left & 73.47 & 56.49 & 2.86  & 95.99 & 57.20  \\ 
    Right & 5.73  & 19.27 & 6.87  & 36.83 & 17.18 & Right & 79.01 & 35.11 & 95.80 & 2.29  & 53.05\\
    \bottomrule
    \end{tabular}
    \caption{The accuracy of each relation type (column-label) for each spatial relation (row-label) presented in the input context for Qwen2-72B and GPT-40 with 0-shot and SG+CoT settings.}
    \label{tab:spatial_lexicon_diagnostic}
\end{table*}

According to Table~\ref{tab:spatial_lexicon_diagnostic}, we observe that Qwen2-72B does not always extract the relation in C-split. However, it shows some evidence of extraction when the spatial lexical form is “front” or “back” as the accuracy is higher in these cases. In scenarios where the spatial lexical form is either “left” or “right,” the model appears to attempt reasoning (i.e., going beyond extracting the same relation mentions in the context), but often fails to produce correct results. For GPT-4o (multimodal), the model demonstrates overall better performance than Qwen2, except in cases that require conversion to “left” or “right.” This may be attributed to errors introduced by facing direction, where the model appears to rely too heavily on reasoning patterns conditioned on the object facing toward the camera. Nevertheless, these results also indicate that the SG+CoT approach provides clear improvements when reasoning is required, though confusion between “left” and “right” persists.

\section{Template Variations Analysis}\label{appendix:templates_experiment}
\subsection{Performance for each question template}
We examine the effect of each question template on the performance of the models.
Table~\ref{tab:template_differece} reports average accuracy on the C-split across all question templates for Qwen2-72B and GPT-4o. 
While results show notable variation across templates, no single template consistently outperforms others across all tasks. Template effectiveness depends on both the prompting technique and the underlying model. Simpler templates generally yield higher accuracy, though more complex expressions may reduce overall performance.
Nevertheless, the comparative trend remains consistent, as SG+CoT often outperforms CoT in overall improvement across both models. Notably, SG prompting yields the smallest performance variation across templates.

\begin{table*}[t]
\centering
\small
\begin{tabular}{ l c c c c c c c }
\toprule
\textbf{Model} & \textbf{T0} & \textbf{T1} & \textbf{T2} & \textbf{T3} & \textbf{T4} & \textbf{Avg.} & \textbf{Highest $\Delta$} \\
\hline
Qwen2 (0-shot)       & 70.23 & 61.36 & 69.35 & 65.55 & 67.97 & 66.91 & 8.87 \\
Qwen2 (4-shot)       & 65.15 & 64.01 & 68.80 & 63.24 & 65.46 & 65.34 & 5.57 \\
Qwen2 (CoT)          & 71.26 & 66.11 & 72.78 & 63.40 & 71.87 & 69.10 & 9.37 \\
Qwen2 (SG + CoT)     & 71.15 & 70.92 & 72.89 & 70.11 & 71.69 & 71.36 & 2.78 \\
\hline
GPT-4o (0-shot)      & 64.02 & 61.14 & 73.16 & 65.77 & 56.66 & 64.19 & 16.50 \\
GPT-4o (4-shot)      & 65.85 & 52.23 & 70.55 & 61.64 & 58.33 & 61.78 & 18.31 \\
GPT-4o (CoT)         & 80.98 & 69.65 & 84.81 & 83.24 & 74.98 & 78.78 & 15.16 \\
GPT-4o (SG + CoT)    & 85.74 & 85.02 & 84.10 & 86.53 & 87.57 & 85.79 & 3.47 \\

\bottomrule
\end{tabular}
\caption{Average accuracy across different templates. Ti refers to the generated QA template $i$ in Table~\ref{tab:templates}. $\Delta$ measures the difference in accuracy between two different templates.}
\label{tab:template_differece}
\end{table*}

\subsection{Order variation in question templates}\label{appendix:inverted_question}
We include an additional template experiment to ensure that results are independent of the order of spatial entities (locatum and relatum) in the questions. A reversed template is shown in Table~\ref{tab:GPT_templates}.

\noindent\textbf{Results.}
According to Table~\ref{tab:A_split-QA_revert} and Table~\ref{tab:QA_c_split_invert},
the average accuracy difference between the original benchmark templates and the reverse versions is relatively small across A-split, C-split (Table~\ref{tab:A_split-QA_revert} and Table~\ref{tab:QA_c_split_invert}). In particular, the difference ranges from 1\% to 4\%, with an average of 2.76\%.
Detailed comparisons can be seen by examining the new C-split of Table~\ref{tab:QA_c_split_invert} alongside Table~\ref{tab:QA_c_split} in the main paper. Specifically, for questions asked from the camera perspective, performance slightly declines when using reverse templates, with the largest drop observed in the 4-shot setting (a decrease of 3.6\%). In contrast, questions asked from the relatum perspective show improvement with reverse templates, particularly in the CoT setting, where accuracy increases by 3.8\%. Notably, the challenge of perspective conversion persists and even worsens with reverse questions, suggesting that these gains are primarily due to cases where the model extracts spatial relations directly from the context without considering perspective. Finally, we observe that Qwen2 exhibits consistent behavior across both question orders by comparing the new results of A-split with Table 1. Qwen2 still favors answering FoR questions with the spatial lexicon explicitly.

\begin{table*}[t]
    \setlength{\tabcolsep}{0.9mm}
    \small
    \centering
    \begin{tabular}{l c c c | c c c | c | c | c | c c c | c c c | c }
    \toprule
     & \multicolumn{9}{c |}{Question with Camera Perspective} & \multicolumn{7}{c}{Question with Relatum Perspective} \\
    \cline{2-17}
     Model & \multicolumn{3}{c|}{Cow} & \multicolumn{3}{c | }{Car} & Box & Pen & Avg. &\multicolumn{3}{c |}{Cow} & \multicolumn{3}{c |}{Car} & Avg. \\
      \cline{2-17}
       & R\% & I\% & Acc. & R\% & I\% & Acc. & Acc. & Acc. & Acc. &R\% & I\% & Acc. & R\% & I\% & Acc. & Acc. \\ 
        \hline
Qwen2-72B (1)& $70.6$ & $29.0$ & $94.1$ & $69.0$ & $30.6$ & $92.6$ & $100.0$ & $100.0$ & $94.7$ & $26.5$ & $73.1$ & $85.5$ & $27.1$ & $72.5$ & $87.6$ & $85.8$ \\
Qwen2-72B (2) & $68.6$ & $31.0$ & $87.7$ & $65.9$ & $33.7$ & $85.1$ & $100.0$ & $100.0$ & $89.1$ & $28.7$ & $70.8$ & $81.7$ & $24.6$ & $74.9$ & $86.3$ & $82.3$\\
Qwen2-72B (3) & $45.3$ & $54.4$ & $86.1$ & $41.6$ & $58.2$ & $85.5$ & $100.0$ & $100.0$ & $88.0$ & $35.1$ & $64.5$ & $87.8$ & $36.6$ & $63.0$ & $85.7$ & $87.5$ \\
Qwen2-72B (4) & $54.5$ & $45.2$ & $88.9$ & $52.4$ & $47.3$ & $90.6$ & $100.0$ & $100.0$ & $90.7$ & $42.3$ & $57.4$ & $78.3$ & $35.6$ & $64.0$ & $82.8$ & $78.9$ \\
\bottomrule
    \end{tabular}
    \caption{QA accuracy in the A-split with Qwen2-72B using templates where the locatum and relatum order is reverse, across all settings. R\% and I\% represent the percentage the model assumes relative or intrinsic FoR for ambiguous expression, explained in Section~\ref{sec:evaluation_setting}. Acc is the accuracy, and Avg is the micro-average of accuracy. (1): 0-shot, (2): 4-shot, (3): CoT, and (4): SG+CoT.}
    \label{tab:A_split-QA_revert}
\end{table*}

\begin{table*}[ht!]
    \small
    \setlength{\tabcolsep}{1mm}
    \centering
    \begin{tabular}{l c c c c c |c c c c c}
    \toprule
     & \multicolumn{5}{c|}{Question with Camera Perspective} & \multicolumn{5}{c}{Question with Relatum Perspective} \\
    \cline{2 - 11}
    Model & ER (CP) & EI (RP) & II (RP) & IR (CP) & Avg. & ER (CP) & EI (RP) & II (RP) & IR (CP) & Avg. \\
    \hline
     Qwen2-72B (0-shot) & $90.6$ & $32.9$ & $31.8$ & $91.1$ & $64.0$ & $26.4$ & $96.8$ & $99.8$ & $24.4$ & $61.7$ \\
     Qwen2-72B (4-shot) & $87.3$ & $33.9$ & $34.4$ & $72.6$ & $61.7$ & $31.0$ & $96.7$ & $99.4$ & $22.7$ & $63.5$ \\
     Qwen2-72B (CoT) & $80.1$ & $54.2$ & $55.7$ & $66.9$ & $67.4$ & $35.7$ & $96.0$ & $97.5$ & $27.0$ & $65.4$ \\
     Qwen2-72B (SG+CoT) & $96.6$ & $41.7$ & $31.2$ & $93.7$ & $70.6$ & $44.0$ & $92.1$ & $94.7$ & $35.0$ & $67.7$ \\
    \bottomrule
    \end{tabular}
    \caption{QA accuracy in the C-split with Qwen2-72B using templates where the locatum and relatum order is reverse, across all settings. ER, EI, II, and IR  denote external relative, external intrinsic, internal intrinsic, and internal relative FoRs. Avg represents the micro-average accuracy. CP refers to context with camera perspective, while RP denotes context with relatum perspective.}
    \label{tab:QA_c_split_invert}
\end{table*}

\section{Human bias in Ambiguous Cases}\label{appendix:human_study}

\begin{lstlisting}[caption={Insturction for collecting human results on QA using A-split of FoREST dataset.}, label={lst:human_instruction}]
"""
Instruction: 
You will be provided with a scene description that describes the spatial relationship between two objects. 
The scene description may include object orientation, such as A is facing toward the camera, to indicate the direction A is facing. 
You will then receive a question asking about the spatial relationship between A and B from either the camera's perspective or an object's perspective. 
Your task is to answer the question based on your understanding of the given spatial relationship.
"""
\end{lstlisting}

\subsection{Experimental setting}
We provided a Google Form with instructions (Listing~\ref{lst:human_instruction}), followed by 150 questions. These were sampled from relatum categories in the A-split: 25 per case, covering four cases (Cow, Car, Box, Pen) for camera-perspective questions, and two cases (Cow, Car) for relatum-perspective questions, as Box and Pen lack intrinsic direction. No time constraints were imposed, and all participants were compensated at the standard research assistant rate. At the end, participants were informed about the study details and asked for consent to use their responses in the analysis. Results were evaluated using the metric defined in Section~\ref{sec:evaluation_setting}, and participant accuracies were averaged to establish the human baseline.

\subsection{Experimental results}

Individual human results are shown in Table~\ref{tab:A_split-QA_human}. These results indicate that humans rely heavily on their assumptions when interpreting ambiguous frame-of-reference (FoR) contexts. Most participants favored the intrinsic FoR in A-split scenarios, though this preference varied across individuals, suggesting that background or prior visual experience may influence FoR interpretation. This aligns with findings from cognitive studies on how humans describe spatial relations, which indicate that recent examples may influence~\cite{cueBySeenAndHear}.
However, our results contrast with other studies suggesting that humans tend to adopt an egocentric (relative) FoR in scenes with fewer landmarks and an allocentric (intrinsic) FoR in those with more landmarks~\cite{egocentricHelp}. Since our study does not provide visual input, we cannot infer how participants mentally reconstruct scenes to answer the questions with only two objects in the scene.
The next notable point is that while Box and Pen cases appear straightforward for models (Table~\ref{tab:A_split-QA}), humans may still struggle with questions in these categories. 
Humans do not exhibit the same limitations as LLMs in perspective-shifting tasks. When the context and question differ in perspective, humans still achieve over 90\% accuracy, whereas LLMs often struggle with this type of reasoning.

\begin{table*}[t]
    \setlength{\tabcolsep}{1mm}
    \small
    \centering
    \begin{tabular}{ l  c c  c | c c  c | c | c | c  | c c  c | c c  c| c }
    \toprule
     & \multicolumn{9}{c|}{Question with Camera Perspective} & \multicolumn{7}{c}{Question with Relatum Perspective} \\
    \cline{2-17}
     Model & \multicolumn{3}{c|}{Cow} & \multicolumn{3}{c|}{Car} & Box & Pen & Avg. &\multicolumn{3}{c|}{Cow} & \multicolumn{3}{c|}{Car} & Avg. \\
      \cline{2-17}
       & R\% & I\% & Acc. & R\% & I\% & Acc. & Acc. & Acc. & Acc. &R\% & I\% & Acc. & R\% & I\% & Acc. & Acc. \\ 
\hline
P \#1 & $22.7$ & $\mathbf{77.3}$ & $88.0$ & $8.3$ & $\mathbf{91.7}$ & $96.0$ & $92.0$ & $96.0$ & $93.0$ & $20.8$ & $\mathbf{79.2}$ & $96.0$ & $4.2$ & $\mathbf{95.8}$ & $96.0$ & $96.0$ \\
P \#2 & $8.7$ & $\mathbf{91.3}$ & $92.0$ & $0.0$ & $\mathbf{100.0}$ & $96.0$ & $36.0$ & $56.0$ & $70.0$ & $20.0$ & $\mathbf{80.0}$ & $100.0$ & $16.7$ & $\mathbf{83.3}$ & $96.0$ & $98.0$ \\
P \#3 & $\mathbf{78.3}$ & $21.7$ & $92.0$ & $\mathbf{75.0}$ & $25.0$ & $96.0$ & $88.0$ & $96.0$ & $93.0$ & $\mathbf{83.3}$ & $16.7$ & $96.0$ & $\mathbf{87.5}$ & $12.5$ & $96.0$ & $96.0$ \\
\hline
Human-baseline & $36.6$ & $\mathbf{63.4}$ & $90.7$ & $27.8$ & $\mathbf{72.2}$ & $96.0$ & $72.0$ & $82.7$ & $85.3$ & $41.4$ & $\mathbf{58.6}$ & $97.3$ & $36.1$ & $\mathbf{63.9}$ & $96.0$ & $96.7$ \\
\bottomrule
    \end{tabular}
    \caption{QA accuracy in the A-Split of human study. R\% and I\% represent the percentage the model assumes relative or intrinsic FoR for ambiguous expression, explained in Section~\ref{sec:evaluation_setting}. Acc is the accuracy, and Avg is the micro-average of accuracy. (1): 0-shot, (2): 4-shot, (3): CoT, and (4): SG+CoT.}
    \label{tab:A_split-QA_human}
\end{table*}

\section{Qualitative Observations}\label{appendix:examples_results}
\begin{figure}[t]
    \centering
    \includegraphics[width=\linewidth]{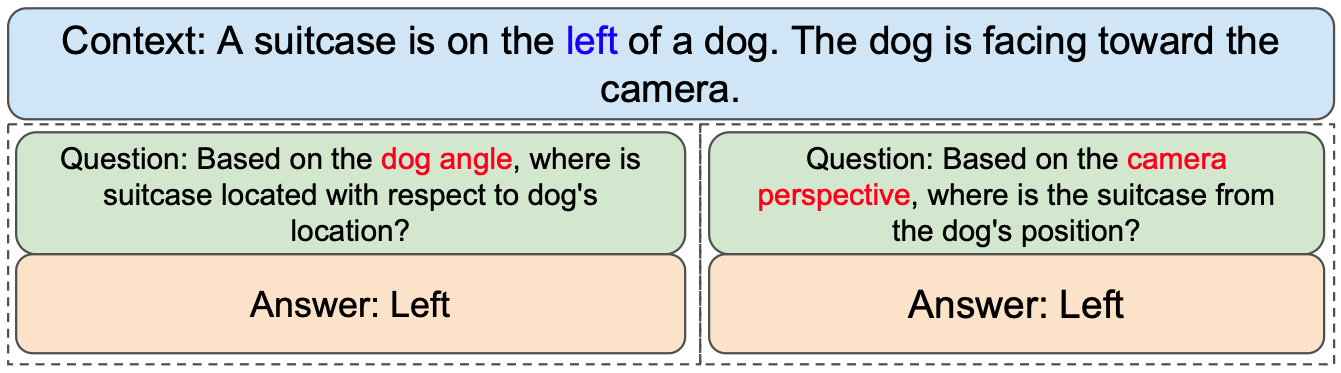}
    \caption{The example in A-split of FoREST, where the model correctly answers both perspectives. In this example, the model responds with the spatial relation in the context, assuming all questions and context have the same perspective.}
    \label{fig:bias1}
\end{figure}

\begin{figure}[t]
    \centering
    \includegraphics[width=\linewidth]{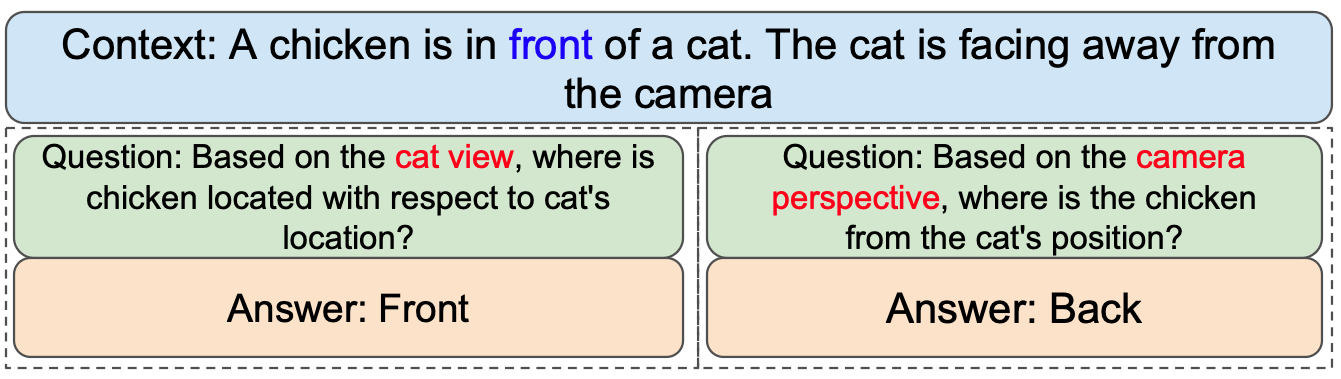}
    \caption{The example in A-split of FoREST, where the model correctly answers both perspectives. In this example, the model assumes the context has a relatum perspective.}
    \label{fig:bias2}
\end{figure}

\begin{figure}[t]
    \centering
    \includegraphics[width=\linewidth]{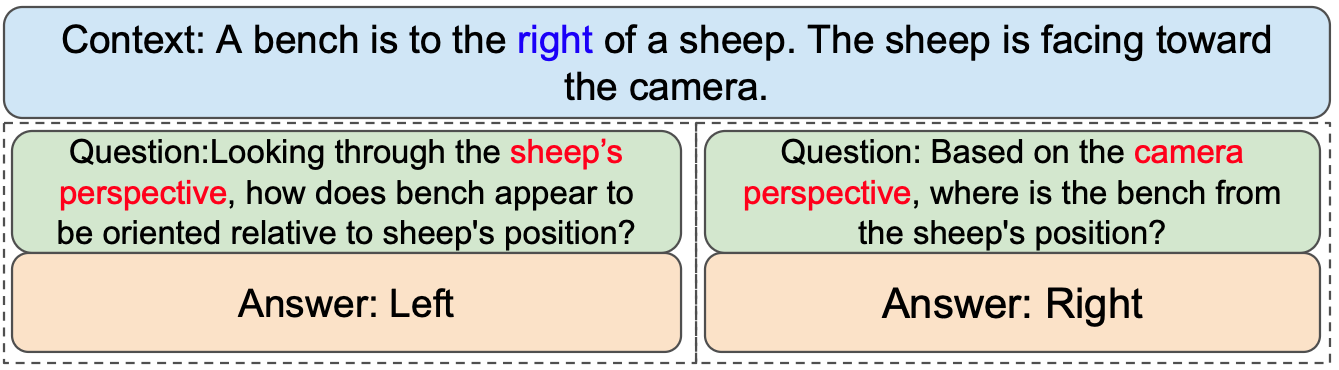}
    \caption{The example in A-split of FoREST, where the model correctly answers both perspectives. In this example, the model assumes the context has a camera perspective.}
    \label{fig:bias3}
\end{figure}

\noindent\textbf{Example of QA Bias in the A-split.}
To illustrate example biases in the A-split, we present three types of model behavior. The first, shown in Figure~\ref{fig:bias1}, occurs when the model assumes a shared perspective between context and question, allowing it to extract spatial relations directly without FoR reasoning. This scenario is the most common behavior of Qwen2. 
The second, illustrated in Figure~\ref{fig:bias2}, arises when the model assumes the context always reflects the relatum’s perspective, favoring intrinsic over relative interpretation.  Lastly, Figure~\ref{fig:bias3} shows the model assuming the context always adopts the camera perspective.

\begin{figure}[t]
    \centering
    \includegraphics[width=\linewidth]{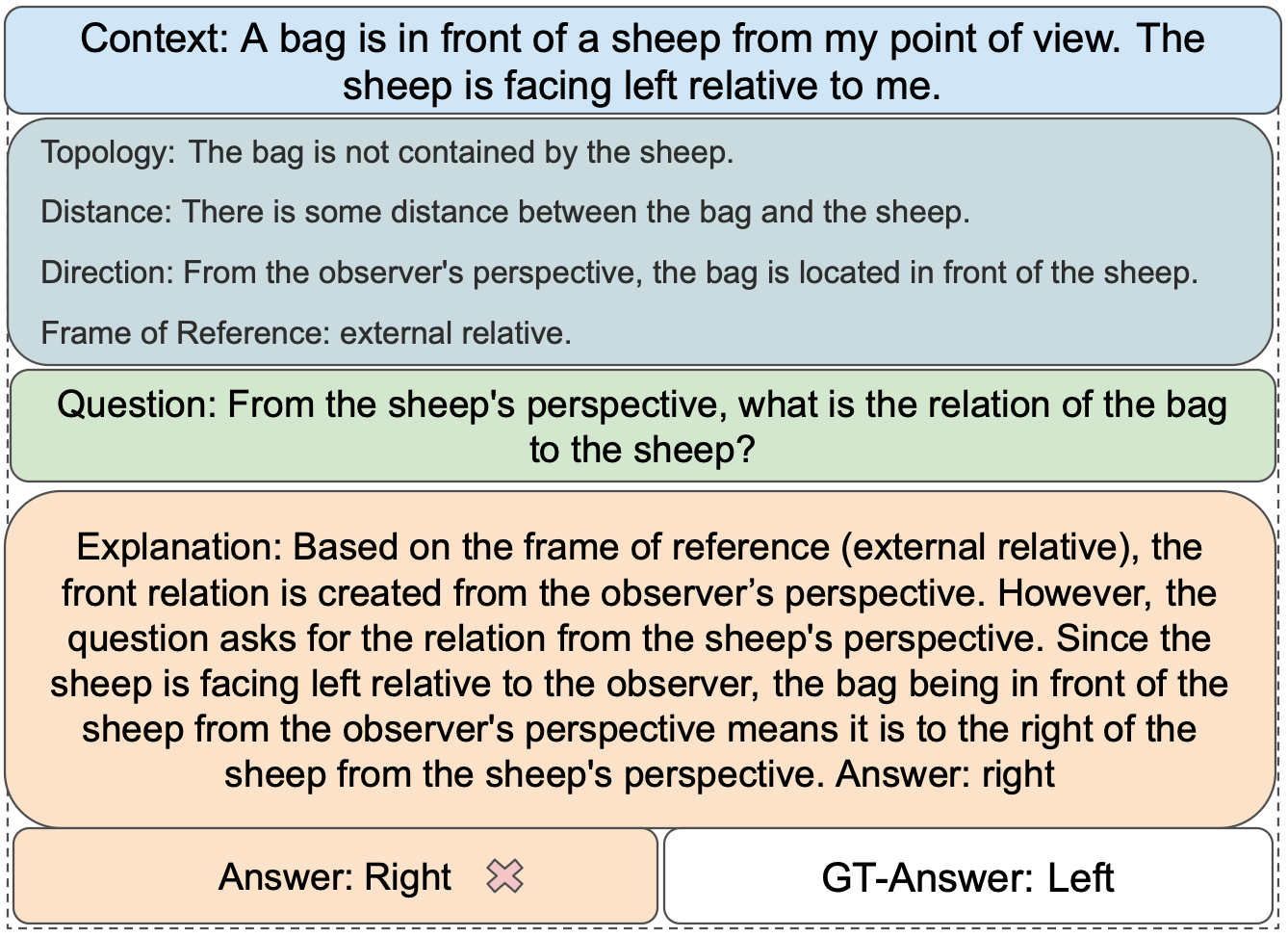}
    \caption{An example of an incorrect SG+CoT case, where the model generates incorrect reasoning and incorrectly converts left to right.}
    \label{fig:C-split-incorrect}
\end{figure}

\noindent\textbf{Incorrect Reasoning in the C-split QA Task.}
To demonstrate LLMs’ confusion in left–right scenarios, we present a quantitative example of a failure case in Figure~\ref{fig:C-split-incorrect}.

\section{VISOR-\textit{uncond} Results}\label{appendix:Visor_uncond}

As mentioned in Section~\ref{sec:evaluation_setting}, VISOR$_{uncond}$ provides the overall spatial relation score, including images with object generation errors. Because it is less focused on evaluating spatial interpretation than VISOR$_{cond}$, which explicitly assesses a text-to-image model’s spatial reasoning, we report VISOR$_{uncond}$ results here in Table~\ref{tab:VISOR_uncode} rather than in the main paper. The results follow a similar pattern with the VISOR$_{cond}$ metric, that is, base models (SD-1.5 and SD-2.1) perform better under the relative frame of reference, while layout-to-image models (e.g., GLIGEN) perform better under the intrinsic frame of reference.

\begin{table*}[t!]
    \centering
    \small
    \begin{tabular}{l  c c  c | c c  c }
    \toprule
         & \multicolumn{6}{c}{VISOR(\%)} \\ \cline{2-7}
        & \multicolumn{3}{c|}{ A-Split } & \multicolumn{3}{c}{ C-Split }  \\ \cline{2 - 7}
        Model & uncond (I) & uncond (R) & uncond (avg) & uncond (I) & uncond (R) & uncond (avg) \\
        \hline

        SD-1.5 & $ 45.43$  & $33.22$ & $ 43.51$ &  $ 35.06$  & $ 35.68$ & $ 35.40$ \\
        SD-2.1 & $\mathbf{62.87}$  & $ 43.90$ & $\mathbf{59.89}$ & $\mathbf{45.98}$  & $ 46.59$ & $\mathbf{46.31}$ \\
        \hline
        Llama3-8B + GLIGEN & $ 46.74$  & $ 38.16$ & $ 45.39$ & $ 33.98$  & $ 39.36$ & $ 36.89$ \\
        Llama3-70B + GLIGEN & $ 54.33$  & $ 46.89$ & $ 53.17$ & $ 38.04$  & $ 46.04$ & $ 42.37$ \\
        Llama3-8B + SG + GLIGEN (Our) & $ 51.83$  & $ 43.24$ & $ 50.48$ & $ 36.28$  & $ 44.43$ & $ 40.70$ \\
        Llama3-70B + SG + GLIGEN (Our) & $ 58.92$  & $\mathbf{47.44}$ & $ 57.12$ & $ 38.23$  & $\mathbf{48.62}$ & $ 43.86$ \\
        \bottomrule
    \end{tabular}
    \caption{VISOR$_{uncond}$ score on the A-Split and C-Split where $I$ refers to the Cow Case and Car Case, where relatum has intrinsic directions, and $R$ refers to the Box Case and Pen case, where relatum lacks intrinsic directions, $avg$  is the micro-average of $I$ and $R$.}
    \label{tab:VISOR_uncode}
\end{table*}

 \section{Analysis of SG-prompting in T2I} \label{appedix:anaylize_SG_improment_t2i}
\begin{table}[t]
    \small
    \centering
    \begin{tabular}{l c c c }
         \toprule
         Model & Layout & Layout$_{cond}$  \\
         \hline
          Llama3-8B  & $85.26$ & $88.84$\\
          Llama3-8B + SG  & $85.04$ & $88.86$  \\
          Llama3-70B  & $88.47$ & $93.16$ \\
          Llama3-70B + SG & $91.95$ & $95.45$ \\
          \bottomrule
    \end{tabular}
    \caption{Layout accuracy where spatial relations are left or right relative to the camera. Layout is evaluated for all generated layouts in C-split while Layout$_{cond}$ uses the same testing examples as VISOR$_{cond}$.}
    \label{tab:layout_results}
\end{table}
To further explain the improvements of SG-prompting in the T2I task, we assess generated bounding boxes in the C-split for left–right relations relative to the camera, since these can be evaluated using bounding boxes alone without depth information. As shown in Table~\ref{tab:layout_results}, SG-prompting improved Llama3-70B’s performance by 3.48\%, while Llama3-8B saw a slight decrease of 0.22\%. This evaluation uses all generated layouts from the C-split, differing from the image subset used for VISOR$_{cond}$ in Table~\ref{tab:I_split}. For consistency, we also report the layout$_{cond}$ score in Table~\ref{tab:layout_results}, which shows that Llama3-8B improves within the same evaluation subset as VISOR$_{cond}$. Overall, incorporating FoR information through SG layout diffusion enables Llama3 to generate better spatial configurations, thereby enhancing image generation performance.

\section{Frame of Reference Identification Task}\label{appendix:FoRIdentification}
We evaluate the LLMs' performance in recognizing the FoR classes from given spatial expressions. 
Each model receives a spatial expression $T$ and outputs one FoR class $FoR$ from the valid set of FoR classes, $For \in $ \{external relative, external intrinsic, internal intrinsic, internal relative\}. All in-context learning examples are in the Appendix~\ref{appendix:in-context}.

\subsection{Experimental setting}
\noindent\textbf{Zero-shot model.} We follow the regular setting of \textit{zero-shot} prompting. 
We called the LLM with the instruction prompt and $T$ to find the corresponding FoR, $F$, of given $T$.

\noindent\textbf{Few-shot model.} We manually craft four spatial expressions for each FoR class. 
To avoid creating bias, each spatial expression is ensured to fit in only one FoR class.
We provide these examples in addition to the instruction as a part of the prompt, followed by $T$ and query $F$ from the LLM.

\noindent\textbf{Chain-of-Thought (CoT) model.}
To create CoT~\citep{wei2023chainofthoughtpromptingelicitsreasoning} examples, we modify the prompt to require reasoning before answering.
Then, we manually crafted reasoning explanations for each example used in the few-shot.
Finally, we call the LLMs, adding modified instructions to updated examples, followed by $T$ and query $F$. 

\noindent\textbf{Spatial-Guided Prompting (SG) model.}
This follows the same SG setting described in Section~\ref{sec:SG_prompting}. We prompt the LLM to extract spatial information from the given $T$, expecting it to return the FoR as part of the SG response. We then extract this FoR to obtain $F$. Unlike QA and T2I tasks, this FoR identification task does not invoke additional CoT for downstream reasoning.

\subsection{Evaluation metrics}
We report the accuracy of the model on the multi-class classification task. Note that the expressions in A-split can have multiple correct answers. Therefore, we consider the prediction correct when it is in one of the valid FoR classes for the given spatial expression. 

\begin{table}[t]
    \small
    \centering
    \setlength{\tabcolsep}{0.9mm}
    \begin{tabular}{ l c c | c c }
    \toprule
    \textbf{Model} & \multicolumn{2}{c|}{inherently clear} & \multicolumn{2}{c}{require template} \\ \cline{2-5}
     & CoT & SG & {CoT} & {SG} \\ 
    \hline
    Llama3-70B  & 19.84 & 44.64 \improve{24.80} & 76.72 & 87.39 \improve{10.67}\\
    Qwen2-72B & 58.20 & 84.22 \improve{26.02} & 88.36 & 93.86 \improve{10.67} \\
    GPT-4o & 12.50 & 29.17 \improve{16.67} & 87.73 & 90.74 \improve{3.01}  \\
    \bottomrule
    \end{tabular}
    \caption{Comparison of CoT and SG prompting in the C-split, separated into inherently clear contexts (without FoR ambiguity) and template-dependent contexts (requiring extra information to resolve ambiguity).}
    \label{tab:model_performance}
\end{table}

\begin{table*}[t]
    \small
    \centering
    \setlength{\tabcolsep}{0.9mm}
    \begin{tabular}{ l  c | c c c c   c }
        \toprule
        & A-split & \multicolumn{5}{c}{C-Split} \\ \cline{3-7}
         Model &  & ER-Split & EI-Split & II-Split & IR-Split & Avg. \\
         \hline
         Gemma2-9B (0-shot) & $94.17$ & $\mathbf{94.24}$ & $35.98$ & $53.91$ & $57.66$  & $60.45$\\
          Gemma2-9B (4-shot) & $59.58$  & $55.89$\worse{38.34} & $72.61$\improve{36.63} & $74.22$\improve{20.31} & $54.44$\worse{3.23} & $64.29$\improve{3.84}\\
         Gemma2-9B (CoT) & $60.49$  & $60.49$\worse{33.74} & $60.54$\improve{24.57} & $87.50$\improve{33.59} & $54.03$\worse{3.63} & $65.64$\improve{5.20}\\
          Gemma2-9B (SG)(Our) & $72.67$ & $65.87$\worse{28.37} & $65.54$\improve{29.57} & $53.12$\worse{0.78} & $\mathbf{95.97}$\improve{38.31} & $70.13$\improve{9.68}\\
         \hline
         llama3-8B (0-shot) & $60.21$ & $32.20$ & $90.11$ & $75.78$ & $0.00$ & $49.52$\\
         llama3-8B (4-shot) & $60.14$ & $47.77$\improve{15.58} & $54.35$\worse{35.76} & $100.00$\improve{24.22} & $41.13$\improve{41.13} & $60.81$\improve{11.29}\\
         llama3-8B (CoT) & $61.32$ & $61.06$\improve{28.86} & $97.28$\improve{7.17} & $100.00$\improve{24.22} & $36.29$\improve{36.29} & $73.66$\improve{24.14}\\
         llama3-8B (SG) (Our) & $62.95$ & $63.29$\improve{31.09} & $94.57$\improve{4.46} & $100.00$\improve{24.22} & $43.55$\improve{43.55} & $75.35$\improve{25.83}\\

         \hline
         llama3-70B (0-shot) & $84.23$ & $74.08$ & $9.57$ & $92.19$ & $68.55$ & $61.10$\\
         llama3-70B (4-shot) & $78.47$ & $81.81$\improve{7.72} & $64.89$\improve{55.33} & $100.00$\improve{7.81} & $75.81$\improve{7.26} & $80.63$\improve{19.53}\\
         llama3-70B (CoT) & $69.11$ & $72.05$\worse{2.03} & $97.07$\improve{87.50} & $100.00$\improve{7.81} & $79.44$\improve{10.89} & $87.14$\improve{26.04}\\
         llama3-70B (SG) (Our) & $76.50$ & $78.21$\improve{4.12} & $97.61$\improve{88.04} & $100.00$\improve{7.81} & $72.18$\improve{3.63} & $87.00$\improve{25.90}\\
         \hline
         Qwen2-7B (0-shot) & $83.64$ & $79.97$ & $59.24$ & $77.34$ & $40.73$ & $64.32$\\
        Qwen2-7B (4-shot) & $61.12$ & $50.52$\worse{29.45} & $65.76$\improve{6.52} & $93.75$\improve{16.41} & $56.05$\improve{15.32} & $66.52$\improve{2.20}\\
        Qwen2-7B (CoT) & $72.12$ & $70.81$\worse{9.16} & $63.80$\improve{4.57} & $99.22$\improve{21.88} & $51.61$\improve{10.89} & $71.36$\improve{7.04}\\
        Qwen2-7B (SG) & $70.61$ & $68.00$\worse{11.98} & $71.20$\improve{11.96} & $88.28$\improve{10.94} & $57.26$\improve{16.53} & $71.18$\improve{6.86}\\
        \hline
        Qwen2-72B (0-shot)& $64.46$ & $62.70$ & $100.00$ & $100.00$ & $39.11$ & $75.45$\\
        Qwen2-72B (4-shot)& $79.12$ & $78.73$\improve{16.03} & $99.35$\worse{0.65} & $87.50$\worse{12.50} & $87.10$\improve{47.98} & $88.17$\improve{12.72}\\
        Qwen2-72B (CoT)& $88.54$ & $88.87$\improve{26.18} & $89.57$\worse{10.43} & $93.75$\worse{6.25} & $83.47$\improve{44.35} & $88.91$\improve{13.46}\\
        Qwen2-72B (SG)& $90.51$ & $90.18$\improve{27.49} & $93.26$\worse{6.74} & $98.44$\worse{1.56} & $85.08$\improve{45.97} & $91.74$\improve{16.29}\\
        \hline
         GPT3.5 (0-shot) & $83.11$ & $88.15$ & $17.50$ & $70.31$ & $41.13$ & $54.27$\\
         GPT3.5 (4-shot) & $61.25$  & $48.95$\worse{39.20} & $62.72$\improve{45.22} & $100.00$\improve{29.69} & $28.63$\worse{12.50} & $60.07$\improve{5.80}\\

         GPT3.5 (CoT) & $66.55$ & $66.62$\worse{21.53} & $96.85$\improve{79.35} & $100.00$\improve{29.69} & $50.81$\improve{9.68} & $78.57$\improve{24.30}\\
         GPT3.5 (SG) (Our) & $70.61$  & $73.30$\worse{14.86} & $92.93$\improve{75.43} & $99.22$\improve{28.91} & $49.19$\improve{8.06} & $78.66$\improve{24.39}\\
         \hline
         GPT4o (0-shot) & $73.82$  & $71.27$ & $98.80$ & $100.00$ & $70.56$ & $85.16$\\
         GPT4o (4-shot) & $66.23$  & $67.87$\worse{3.40} & $98.70$\worse{0.11} & $100.00$\improve{0.00} & $78.63$\improve{8.06} & $86.30$\improve{1.14}\\
         GPT4o (CoT) & $72.44$  & $72.77$\improve{1.51} & $100.00$\improve{1.20} & $100.00$\improve{0.00} & $73.79$\improve{3.23} & $86.64$\improve{1.48}\\
         GPT4o (SG) (Our) & $76.44$ & $74.67$\improve{3.40} & $97.72$\worse{1.09} & $100.00$\improve{0.00} & $68.55$\worse{2.02} & $85.23$\improve{0.08}\\
         \bottomrule
    \end{tabular}
    \caption{Accuracy results report from FoR Identification with LLMs. The correct prediction is one of the valid FoR classes for the given spatial expression. All FoR classes are external relative (ER), external intrinsic (EI), internal intrinsic (II), and internal relative (IR).}
    \label{tab:text_experiment}
\end{table*}

\begin{figure*}[t]
    \centering
    \begin{subfigure}[ht]{0.49\textwidth}
        \centering
        \includegraphics[width=0.88\textwidth, trim={0 0 0 2cm}]{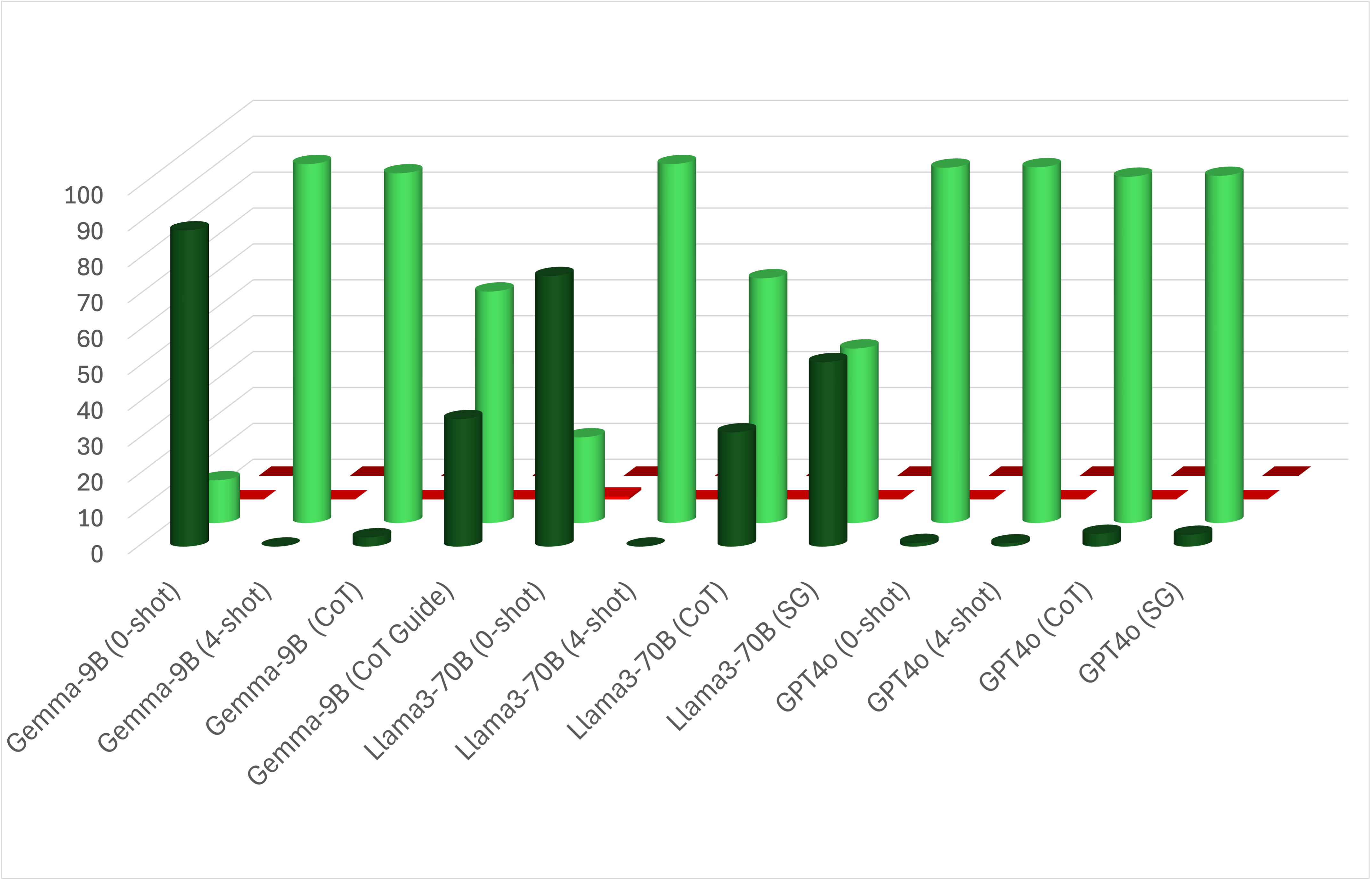}
        \caption{Results of Cow Case in A-Split. 
        }
    \end{subfigure}%
    ~ 
    \begin{subfigure}[ht]{0.49\textwidth}
        \centering
        \includegraphics[width=0.88\textwidth, trim={0 0 0 2cm}]{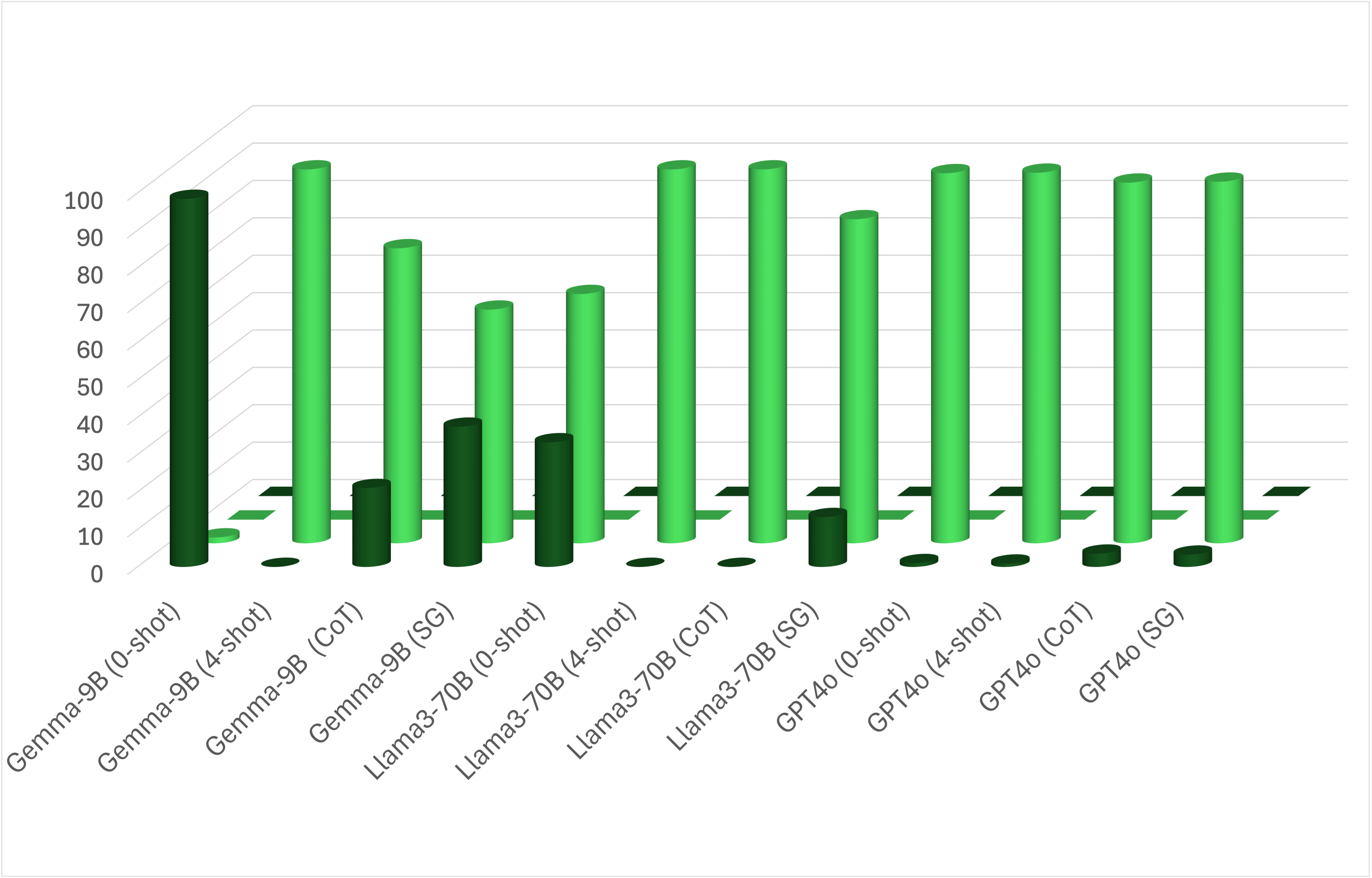}
        \caption{Results of Car Case in A-Split. 
        }
    \end{subfigure}
    
    \vskip\baselineskip
    
    \begin{subfigure}[ht]{0.49\textwidth}   
        \centering 
        \includegraphics[width=0.88\textwidth, trim={0 0 0 1cm}]{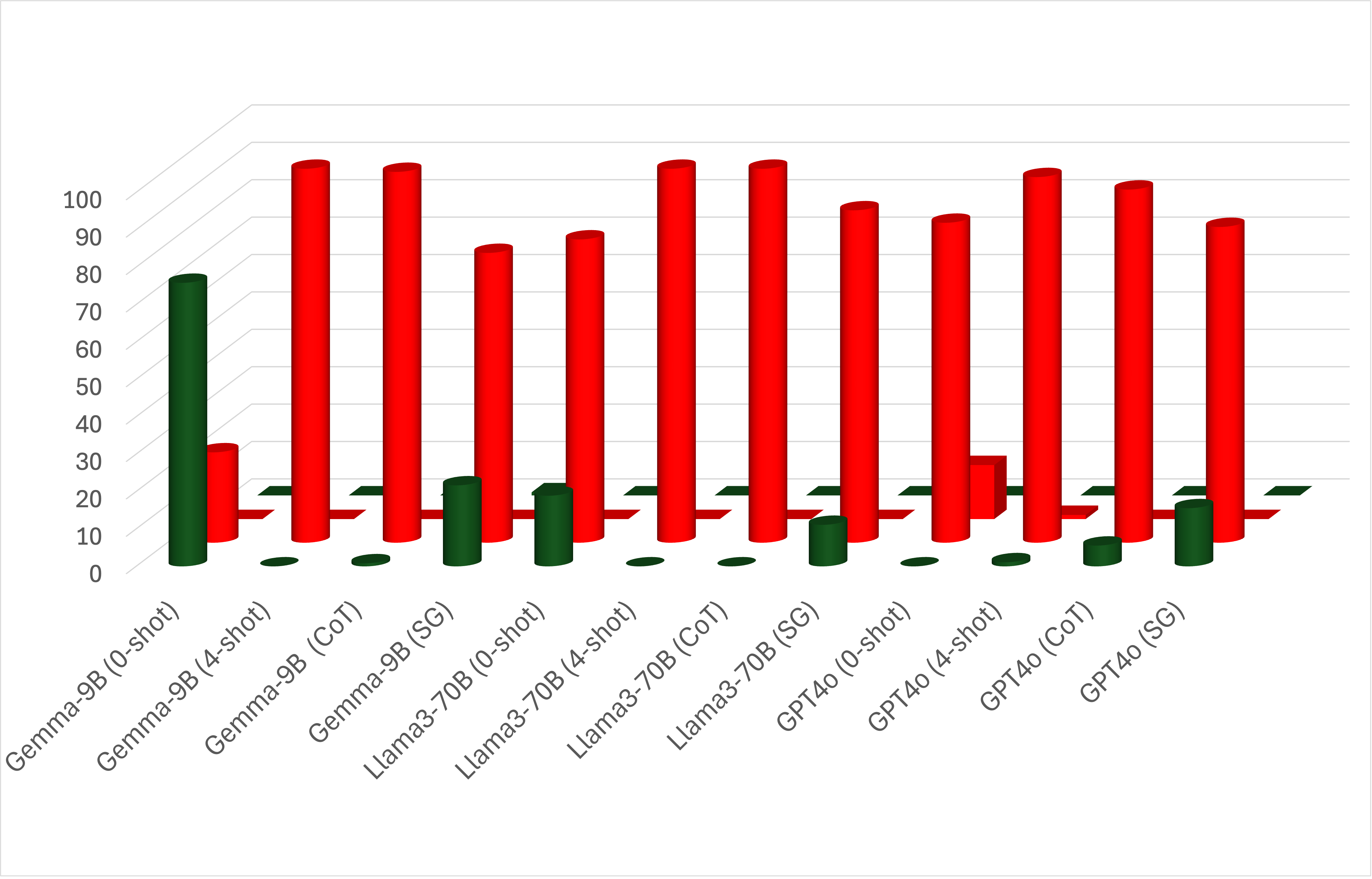}
        \caption{Results of Box Case in A-Split. 
        }    
    \end{subfigure}
        ~
    \begin{subfigure}[ht]{0.49\textwidth}   
        \centering 
        \includegraphics[width=0.88\textwidth, trim={0 0 0 1cm}]{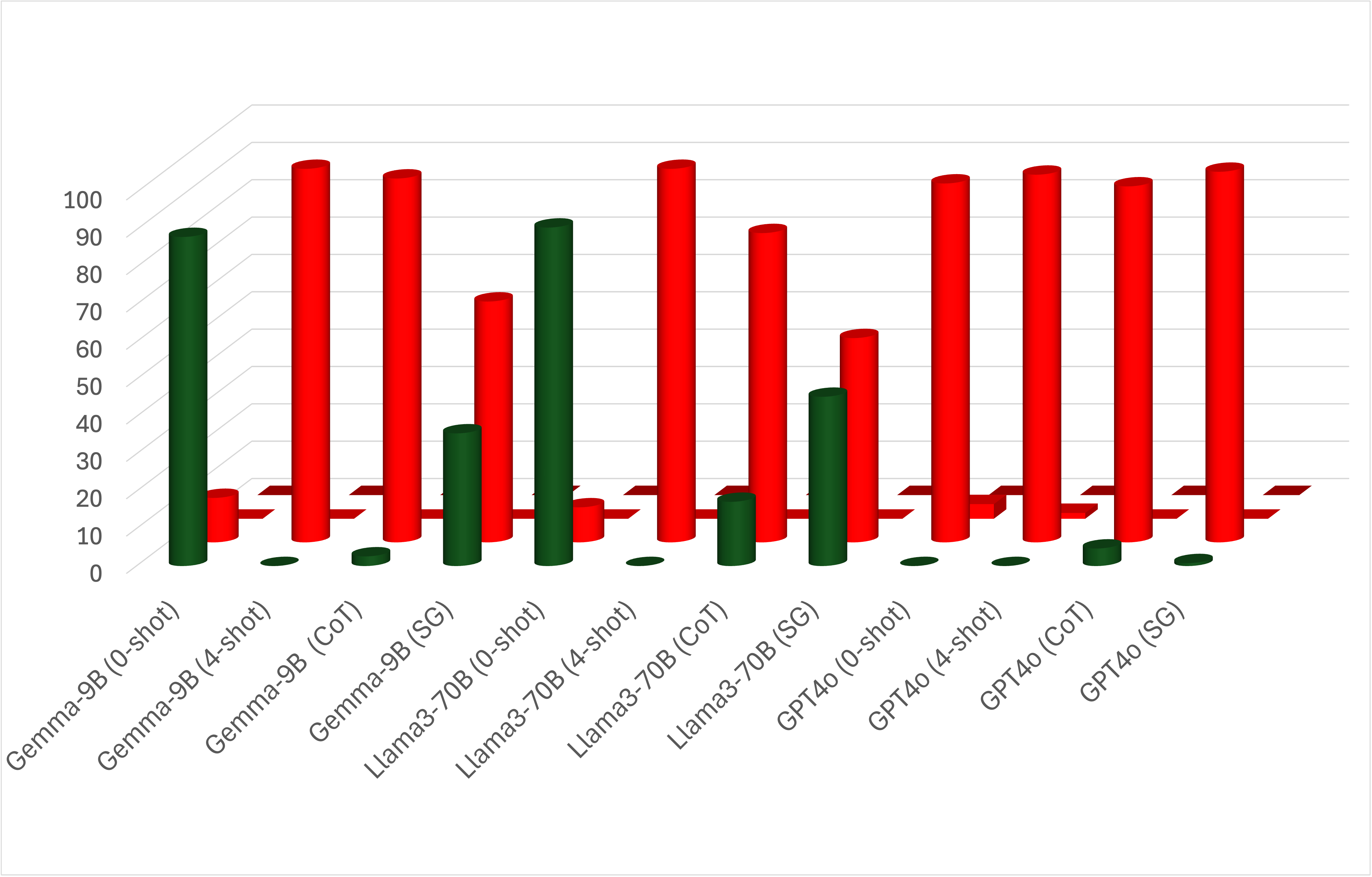}
        \caption{Results of Pen Case in A-Split. 
        }  
    \end{subfigure}
    \caption{Red indicates incorrect FoR identifications and green indicates correct ones. Dark colors represent relative FoRs, while light colors represent intrinsic FoRs. Round shapes correspond to external FoRs, and squares correspond to internal FoRs. The plot depth represents the four FoRs—external relative, external intrinsic, internal intrinsic, and internal relative—from front to back. This plot shows the results for Gemma-9B, Llama3-72B, and GPT4o.}
    \label{fig:cow_car_case}
\end{figure*}

\begin{figure*}[t]
    \centering
    \begin{subfigure}[t]{0.49\textwidth}
        \centering
        \includegraphics[width=0.88\textwidth, trim={0 0 0 0}]{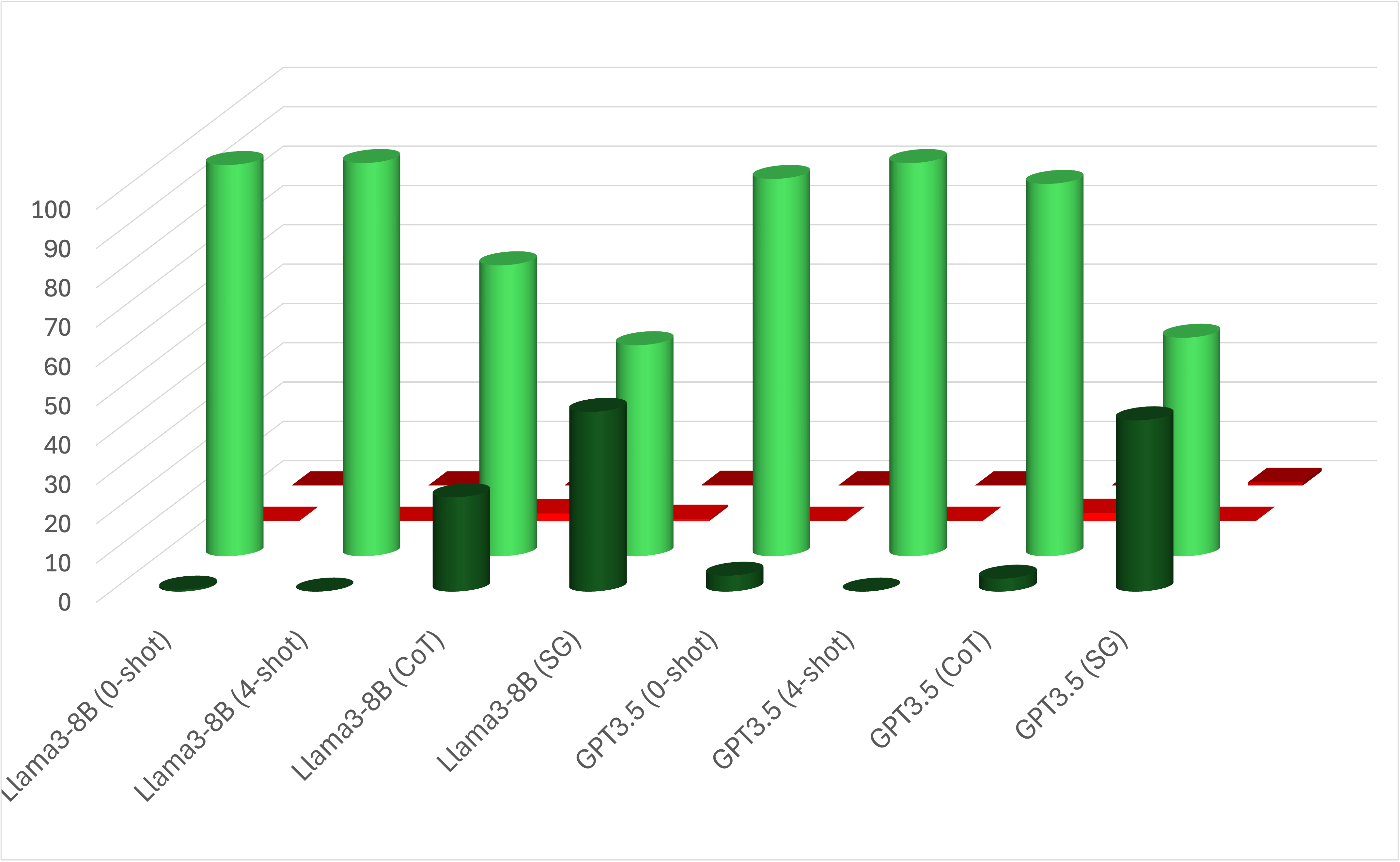}
        \caption{Results of Cow Case in A-Split. 
        }
    \end{subfigure}%
    ~ 
    \begin{subfigure}[t]{0.49\textwidth}
        \centering
        \includegraphics[width=0.88\textwidth, trim={0 0 0 0}]{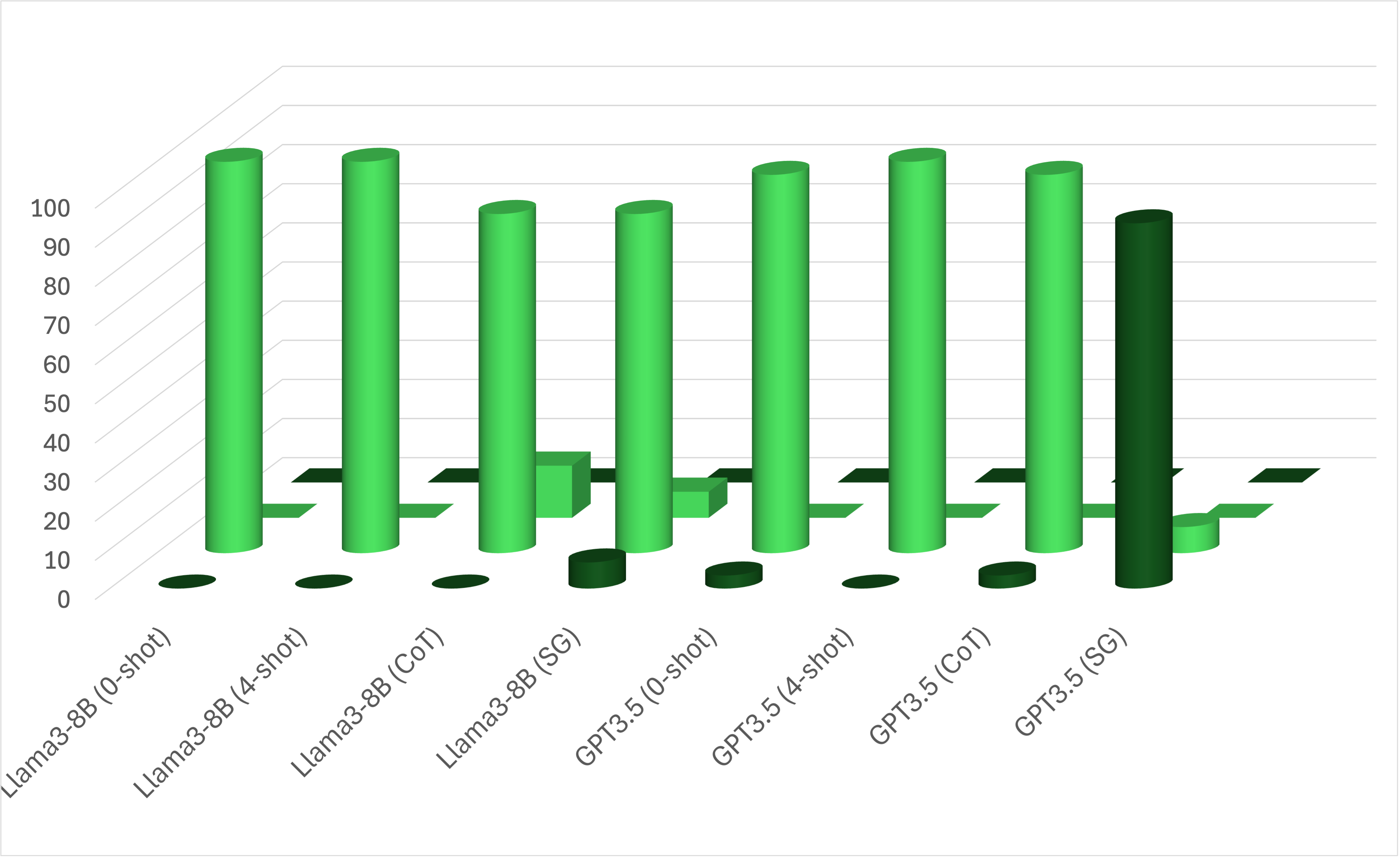}
        \caption{Results of Car Case in A-Split. 
        }
    \end{subfigure}
    
    \vskip\baselineskip
    
    \begin{subfigure}[t]{0.49\textwidth}   
        \centering 
        \includegraphics[width=0.88\textwidth, trim={0 0 0 0}]{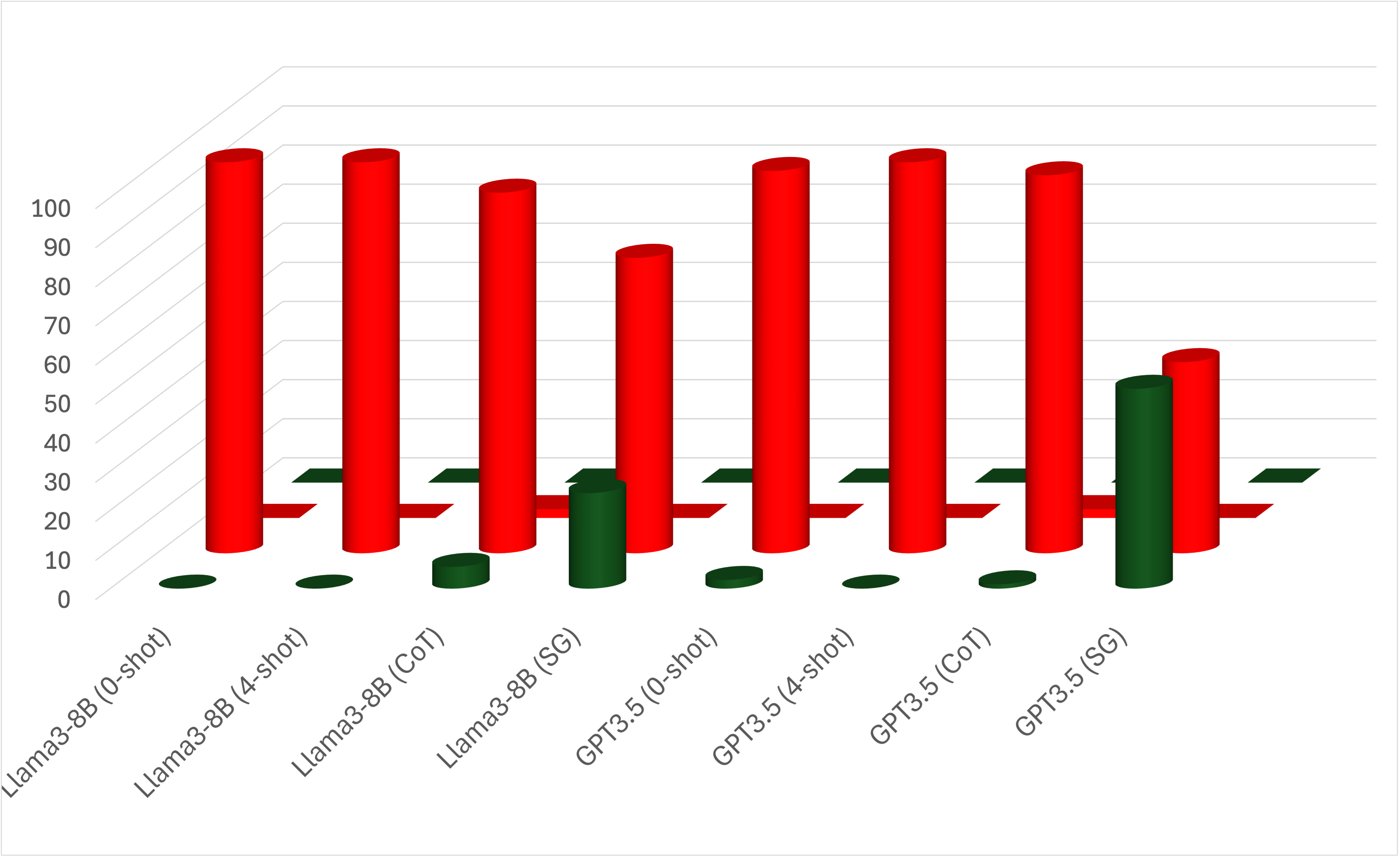}
        \caption{Results of Box Case in A-Split. 
        }    
    \end{subfigure}
        ~
    \begin{subfigure}[t]{0.49\textwidth}   
        \centering 
        \includegraphics[width=0.88\textwidth, trim={0 0 0 0}]{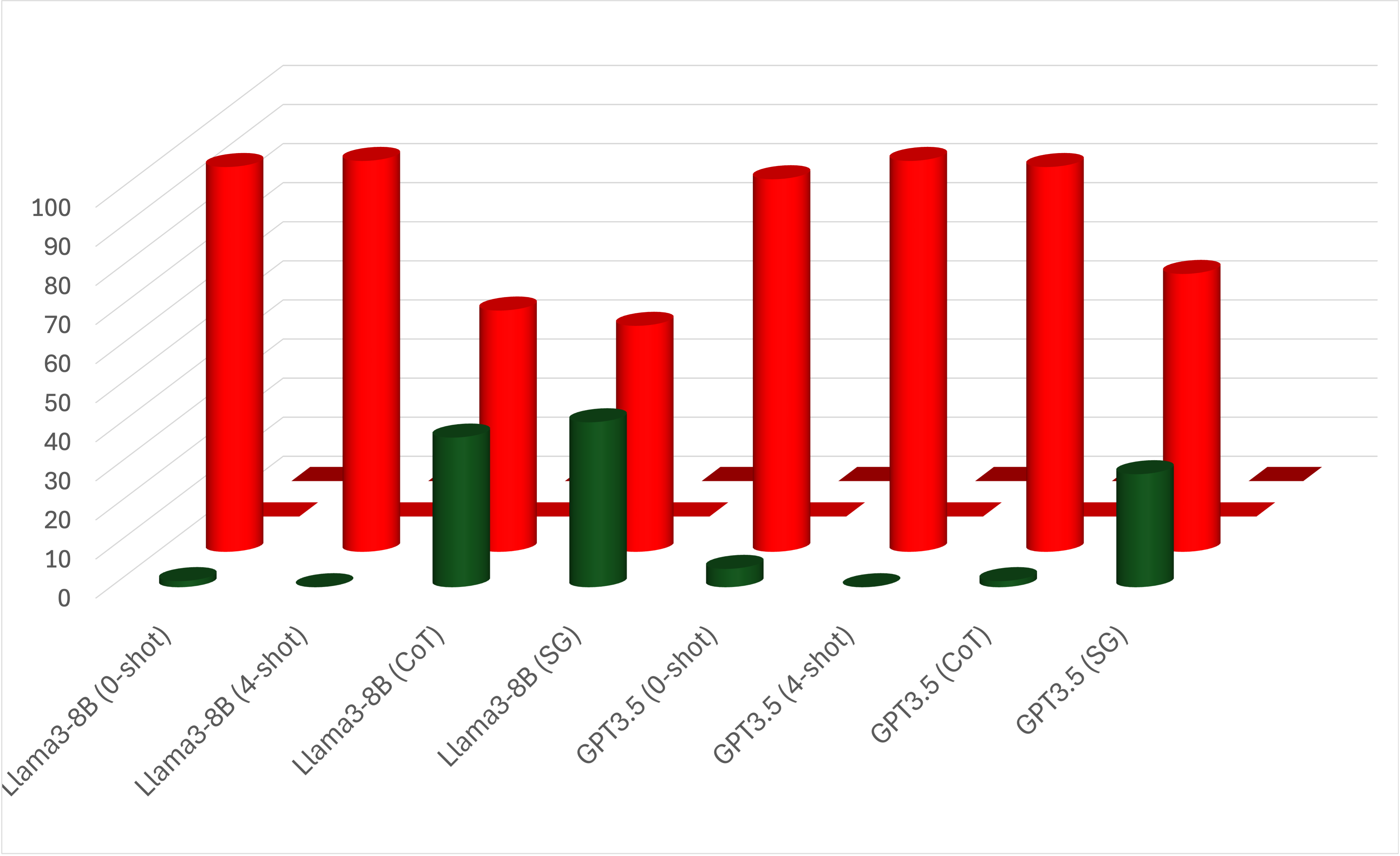}
        \caption{Results of Pen Case in A-Split. 
        }  
    \end{subfigure}
    \caption{Red indicates incorrect FoR identifications and green indicates correct ones. Dark colors represent relative FoRs, while light colors represent intrinsic FoRs. Round shapes correspond to external FoRs, and squares correspond to internal FoRs. The plot depth represents the four FoRs—external relative, external intrinsic, internal intrinsic, and internal relative—from front to back. This plot shows the results for Llama3-8B and GPT3.5.}
    \label{fig:cow_car_case2}
\end{figure*}

\subsection{Experimental results}

\subsubsection{Inherent FoR bias in LLMs} 
\noindent\textbf{C-spilt.}
The \textit{zero-shot} setting reflects the LLMs’ inherent bias in identifying FoR. Table~\ref{tab:text_experiment} shows accuracy for each FoR class in the C-split, where sentences explicitly include topology and perspective information. Some models strongly favor specific FoR classes: notably, Gemma2-9B achieves near-perfect accuracy on external relative FoR but performs poorly on others—especially external intrinsic—indicating a strong bias toward external relative. In contrast, GPT-4o and Qwen2-72B perform well on intrinsic FoR classes but poorly on relative ones.

\noindent\textbf{A-spilt.}
We examine FoR bias in the A-split. Based on the results in Table~\ref{tab:text_experiment}, we plot the top three models (Gemma2-9B, Llama3-70B, and GPT-4o) for detailed analysis in Figures~\ref{fig:cow_car_case}. The plots show the frequency distribution of FoR categories. Gemma2-9B and GPT-4o display strong biases toward external relative and external intrinsic, respectively. This bias benefits Gemma2-9B in the A-split, since all spatial expressions can be interpreted as external relative. By contrast, GPT-4o’s bias leads to errors when intrinsic FoRs are invalid, as in the Box and Pen cases (plots (c) and (d)).
Llama3 shows a different pattern, with its bias depending on the relatum’s properties, particularly the container affordance. In cases where the relatum cannot serve as a container (Cow and Pen), Llama3 favors the external relative. Conversely, when the relatum has container potential, Llama3 tends to favor external intrinsic.

\subsubsection{Effect of ICL variations}\label{sec:result_A_ICL}

\noindent\textbf{C-spilt.}
We evaluate model behavior under different in-context learning (ICL) methods. As shown in Table~\ref{tab:text_experiment}, \textit{few-shot} prompting improves performance over \textit{zero-shot} across multiple LLMs by reducing their bias toward specific classes, though this reduction sometimes lowers performance (e.g., Gemma2 in the external relative). Another observation is that while \textit{CoT} generally improves performance in larger LLMs, it can be counterproductive in smaller models for some FoR classes, likely because they struggle to infer FoR from longer contexts. A similar negative effect appears in SG prompting, which also uses longer explanations.
Despite these degradations in small models, SG prompting performs well across architectures and achieves outstanding results with Qwen2-72B. To better understand this, we compare CoT and SG prompting in Table~\ref{tab:model_performance}. CoT shows large performance gaps between contexts with inherently clear FoR and those requiring templates to resolve ambiguity, indicating its reliance on template-specific cues. By contrast, SG prompting exhibits a smaller gap and substantially outperforms CoT in inherently clear contexts. This suggests that guiding models to identify topological, distance, and directional relation characteristics enhances FoR comprehension.

\noindent\textbf{A-spilt.}
We use Figure~\ref{fig:cow_car_case} to analyze behavior under ICL of the A-split. The A-split shows minimal improvement overall, though some notable changes emerge. With \textit{few-shot}, all models shift toward external intrinsic FoR—even when the relatum lacks intrinsic direction (Box and Pen)—a bias also observed in Gemma2-9B, which usually behaves differently. This indicates that models inherit biases from examples despite efforts to avoid them.
\textit{CoT} reduces some of this bias, encouraging LLMs to predict relative FoR, which is generally valid across scenarios. For example, Gemma2 predicts relative FoR for Cow and Car, while Llama3 does so for Cow and Pen, where the relatum cannot act as a container. GPT-4o shows slight improvements across all cases without relying on relatum properties.
Unlike \textit{CoT}, SG prompting is effective across scenarios, significantly reducing biases while following a similar adjustment pattern. Specifically, it increases external relative predictions for Car and Cow in Gemma2-9B and for Cow and Pen in Llama3-70B. GPT-4o shows only slight bias reduction, yet overall performance improves for most models (Table~\ref{tab:text_experiment}). Llama3-70B’s behavior is mirrored in Llama3-8B and GPT-3.5, with corresponding plots shown in Figure~\ref{fig:cow_car_case2}.

\subsubsection{Experiment with different temperatures}
We conducted additional experiments to investigate the impact of temperature on model bias in the A-split of the FoREST dataset. As shown in Table~\ref{tab:temp_table}, comparing temperatures 0 and 1 revealed distribution shifts of up to 10\%. However, the relative preferences across most categories remain unchanged. In particular, the model produced the highest-frequency responses for the Cow, Car, and Pen cases, with some increases under certain settings. Overall, higher temperature does not substantially increase the diversity of LLM responses in this task, which is a notable finding.

\begin{table*}[t]
    \small
    \centering
    \begin{tabular}{l c c | c c | c c | c c }
    \toprule
    Model & \multicolumn{2}{c}{ER} & \multicolumn{2}{|c}{EI} & \multicolumn{2}{|c}{II} & \multicolumn{2}{|c}{IR} \\
    & temp-0 & temp-1 & temp-0 & temp-1 & temp-0 & temp-1 & temp-0 & temp-1 \\
    \midrule
    \multicolumn{9}{ c }{Cow Case} \\
    \midrule
     0-shot   & 75.38 &  87.12 & 23.86 & 12.50 & 0.76 & 0.13 & 0.00 & 0.25\\ 
     4-shot   & 0.00 &  15.66 & 100.00 & 84.34 & 0.00 & 0.00 & 0.00 & 0.00\\
     CoT & 31.82 & 49.87 & 68.18 & 49.87 & 0.00 & 0.13 & 0.00 & 0.13 \\
     SG & 51.39 & 70.45 & 48.61 & 29.42 & 0.00 & 0.00 & 0.00 & 0.13\\
     \midrule
     \multicolumn{9}{ c }{Box Case} \\
     \midrule
     0-shot   & 22.50 &  41.67 & 77.50 & 58.33 & 0.00 & 0.13 & 0.00 & 0.25\\ 
     4-shot   & 0.00 &  0.00 & 100.00 & 100.00 & 0.00 & 0.00 & 0.00 & 0.00\\
     CoT & 0.00 &  5.83 & 100.00 & 94.17 & 0.00 & 0.00 & 0.00 & 0.00\\
     SG & 11.67 &  33.33 & 88.33 & 66.67 & 0.00 & 0.00 & 0.00 & 0.00\\
     \midrule
     \multicolumn{9}{ c }{Car Case} \\
     \midrule
     0-shot   & 55.20 & 68.24 & 49.01 & 31.15 & 0.79 & 0.61 & 0.00 & 0.00\\ 
     4-shot   & 0.60 &  5.94 & 99.40 & 94.06 & 0.00 & 0.00 & 0.00 & 0.00\\
     CoT & 19.64 &  38.52 & 80.16 & 61.27 & 0.20 & 0.20 & 0.00 & 0.00\\
     SG & 44.25 &  56.97 & 55.75 & 43.03 & 0.00 & 0.00 & 0.00 & 0.00\\
     \midrule
     \multicolumn{9}{ c }{Pen Case} \\
     \midrule
     0-shot   & 90.62 & 96.88 & 9.38 & 3.12 & 0.00 & 0.61 & 0.00 & 0.00\\ 
     4-shot   & 0.00  &  7.03 & 100.00 & 92.97 & 0.00 & 0.00 & 0.00 & 0.00\\
     CoT & 17.19 &  28.91 & 82.81 & 71.09 & 0.20 & 0.20 & 0.00 & 0.00\\
     SG & 48.31 &  57.81 & 54.69 & 42.19 & 0.00 & 0.00 & 0.00 & 0.00\\
     \bottomrule
    \end{tabular}
    \caption{Percentage distribution of responses from Llama3-70B at two different temperatures (0 and 1) on the A-split of FoREST. All FoR classes are external relative (ER), external intrinsic (EI), internal intrinsic (II), and internal relative (IR).}
    \label{tab:temp_table}
\end{table*}

\section{Prompt Specifications}\label{appendix:in-context}
\subsection{FoR identification task}
We provide the prompting for each in-context learning. The prompting for \textit{zero-shot} and \textit{few-shot} is provided in Listing~\ref{lst:base_instruction}. The instruction answer for these two in-context learning is ``Answer only the category without any explanation. The answer should be in the form of \{Answer: Category.\}"

For the Chain of Thought (CoT), we only modified the instruction answer to ``Answer only the category with an explanation. The answer should be in the form of \{Explanation: Explanation Answer: Category.\}" 
Similarly to CoT, we only modified the instruction answer to ``Answer only the category with an explanation regarding topological, distance, and direction aspects. The answer should be in the form of \{Explanation: Explanation Answer: Category.\}", respectively. The example responses are provided in Listing~\ref{lst:example_answer} for Spatial Guided prompting.

\begin{lstlisting}[caption={Prompt for finding the frame of reference class of given context.}, label={lst:base_instruction}]
# Instruction to find frame of reference class of given context
"""
Instruction: 
You specialize in language and spatial relations, specifically in the frame of context (multiple perspectives in the spatial relation). Identify the frame of reference category given the following context. There are four classes of the frame of reference (external intrinsic, internal intrinsic, external relative, internal relative). Note that the intrinsic direction refers to whether the model has the front/back by itself. (Example: a bird, human. Counter Example: a ball, a box). "

External intrinsic. The spatial description of an entity A relative to another entity B, where (1) A is not contained by B, (2) the spatial relation is based on B's facing orientation (intrinsic direction) if B has one.

Internal intrinsic. The spatial description of an entity A relative to another entity B, where (1) A is contained by B, (2) the spatial relation is based on B's facing orientation (intrinsic direction) if B has one.

External relative. The spatial description of an entity A relative to another entity B, where (1) A is not contained by B, (2) The spatial relation is relative to entity C, where C is not involved in the spatial relation and only provides the perspective. An example of C is the camera.

Internal relative. The spatial description of an entity A relative to another entity B, where (1) A is contained by B, (2) The spatial relation is relative to entity C, where C is not involved in the spatial relation and only provides the perspective. An example of C is the camera.

{Instruction answer}
# Normal Instruction answer: Find the frame of reference category without any explanations. The answer should be in the form of Answer: Category.
# COT Instruction answer: Find the frame of reference category with an explanation. The answer should be in the form of Explanation: Explanation. Answer: Category.
# SG Instruction answer: Find the frame of reference category with an explanation. Additionally, include three more explanations relating to topology, distance, and direction from the given context. The answer should be in the form of \"Explanation: Explanation. Topology: Explanation. Distance: Explanation. Direction: Explanation. Answer: Category.

Context: {spatial exprssion}
"""

\end{lstlisting}

\begin{lstlisting}[caption={Spatial expression examples using for few-shots, Chain-of-Thought (CoT), and Spatial Guide (SG) prompting}, label={lst:example_answerQA}]
## These are four examples used in few-shot, CoT, and SG
# Example 1: The bird is outside the car and in front of the car relative to the car.
# Few-shot response: Answer: external intrinsic
# CoT response: Explanation: The car does not contain the bird, and the spatial relation is relative to the car's facing orientation. The intrinsic direction of the car is relevant in this context, as the description \"in front of\" implies a direction relative to the sheep's front or back. Answer: external intrinsic.
# SG response: Explanation: The car does not contain the bird, and the spatial relation is relative to the car's facing orientation. The intrinsic direction of the car is relevant in this context, as the description \"in front of\" implies a direction relative to the sheep's front or back. Topology: The bird is outside the car. Distance: There is little distance between the bird and the car. Direction: The bird is in the front direction of the car (entity B). Answer: external intrinsic.

# Example 2: The bird is inside of the car and left of the car from the car's perspective.
# Few-shot response: Answer: internal intrinsic
# CoT response: Explanation: In this scenario, the bird is contained within the car, meaning it is an internal relationship. Additionally, the description specifies the bird's position relative to the car, indicating that the spatial relation is relative to the car itself, considering its intrinsic direction. Answer: internal intrinsic.
# SG response:  Explanation: In this scenario, the bird is contained within the car, meaning it is an internal relationship. Additionally, the description specifies the bird's position relative to the car, indicating that the spatial relation is relative to the car itself, considering its intrinsic direction. Topological: The bird is inside the car. Distance: The bird is in the same location as the car. Direction: The bird is in the left direction of the car (entity B). Answer: internal intrinsic.

# Example 3: The box is inside and at the back of the room.
# Few-shot response: Answer: internal relative
# CoT response: Explanation: In this context, the box (entity A) is described as inside the room (entity B), fulfilling the first condition for an internal frame of reference. However, the spatial relation given (\" at the back\") cannot originate from the room since the room does not have intrinsic direction, implying it is relative to an external observer or reference point (C), such as a camera or a person viewing the scene. Answer: internal relative
# SG response:  Explanation: In this context, the box (entity A) is described as inside the room (entity B), fulfilling the first condition for an internal frame of reference. However, the spatial relation given (\" at the back\") cannot originate from the room since the room does not have intrinsic direction, implying it is relative to an external observer or reference point (C), such as a camera or a person viewing the scene. Topological: The box is inside the room. Distance: The box is in the same location as the room. Direction: From the observer's perspective, the box is located at the back of the room. Answer: internal relative

# Example 4: A phone is on the left of a tablet from my perspective.
# Few-shot response: Answer: external relative
# CoT response: Explanation: In this context, the spatial relation is described from the perspective of an observer (C) who is not involved in the spatial relation. The phone (A) is not contained by the tablet (B), and the spatial relation \"left of\" is relative to the observer's perspective, not the tablet's facing orientation. Answer: external relative.
# SG response:  Explanation: In this context, the spatial relation is described from the perspective of an observer (C) who is not involved in the spatial relation. The phone (A) is not contained by the tablet (B), and the spatial relation \"left of\" is relative to the observer's perspective, not the tablet's facing orientation. Topological: The phone is not contained by the tablet. Distance: There is some distance between the phone and the tablet. Direction: From the observer's perspective, the phone is located to the left of the tablet. Answer: external relative.
\end{lstlisting}

\subsection{Question answering task}\label{appendix:QA_example}

\begin{lstlisting}[caption={Spatial expression examples using for few-shots, Chain-of-Thought (CoT), and Spatial Guide (SG) prompting for question-answering.}, label={lst:example_answerQA}]
'''
#Context: The bird is outside the car and in front of the car relative to the car. The car is facing toward the camera. 
#Question: Based on the camera's perspective, where is the bird from the car position in the scene? 
#SG information (used in SG + CoT): Frame of reference Explanation: The car does not contain the bird, and the spatial relation is relative to the car's facing orientation. The intrinsic direction of the car is relevant in this context, as the description \"in front of\" implies a direction relative to the sheep's front or back. Topology: The bird is outside the car. Distance: There is little distance between the bird and the car. Direction: The bird is in the front direction of the car (entity B). Frame of reference: external intrinsic.
#Normal response: Answer: front
#CoT / (SG + COT) response:Explanation: Based on the context, the bird's position is in the front direction of the car. The car is facing toward the camera. Then, the car's front direction is the camera's front direction. Therefore, the bird's position is in front of the car's position from the camera's perspective. Answer: front

#Context: The bird is inside the car and left of the car from the car's perspective. The car is facing to the right relative to the camera.  #Question: Based on the camera's perspective, where is the bird from the car's position?
#SG information (used in SG + CoT):  Frame of reference Explanation: In this scenario, the bird is contained within the car, meaning it is an internal relationship. Additionally, the description specifies the bird's position relative to the car, indicating that the spatial relation is relative to the car itself, considering its intrinsic direction. Topological: The bird is inside the car. Distance: The bird is in the same location as the car. Direction: The bird is in the left direction of the car (entity B). Frame of reference: internal intrinsic.
#Normal response: Answer: back
#CoT / (SG + COT) response: Explanation: Based on the context, the bird's position is in the left direction of the car. The car is facing to the right. Then, the car's left direction is the camera's back direction. Therefore, the bird's position is to the back of the car's position from the camera's perspective. Answer: back

Context: The box is inside and to the right of the room from the observer's perspective. 
Question: From the observer's perspective, what is the spatial relation of the box to the room? 
#SG information (used in SG + CoT): Frame of reference Explanation: In this context, the box (entity A) is described as inside the room (entity B), fulfilling the first condition for an internal frame of reference. However, the spatial relation given (\" at the back\") cannot originate from the room since the room does not have intrinsic direction, implying it is relative to an external observer or reference point (C), such as a camera or a person viewing the scene. Topological: The box is inside the room. Distance: The box is in the same location as the room. Direction: From the observer's perspective, the box is located at the back of the room. Frame of reference: internal relative.
#Normal response: Answer: right
#CoT / (SG + COT) response: Explanation: Based on the context, the box is to the right of the room from the camera's direction. Therefore, the box's position is to the right of the room's position from the observer's perspective. Answer: right

Context: A phone is to the left of a tablet from my perspective. The tablet is facing to the right. Question: From my perspective, what is the spatial relation of the phone to the tablet?
#SG information (used in SG + CoT): Frame of Reference Explanation: In this context, the spatial relation is described from the perspective of an observer (C) who is not involved in the spatial relation. The phone (A) is not contained by the tablet (B), and the spatial relation \"left of\" is relative to the observer's perspective, not the tablet's facing orientation. Topological: The phone is not contained by the tablet. Distance: There is some distance between the phone and the tablet. Direction: From the observer's perspective, the phone is located to the left of the tablet. Frame of Reference: external relative.
#Normal response: Answer: left
#CoT / (SG + COT) response: Explanation: Based on the context, the phone is to the left of the tablet from my perspective. The direction of the tablet is not relevant in this context since the left relation is from my perspective. Therefore, from my perspective, the phone is to the left of the tablet. Answer: left
'''
\end{lstlisting}

\subsection{Text to Layout Task}
\begin{lstlisting}[caption={Prompt for generating bounding coordinates to use as the layout for layout-to-image models.}, label={lst:example_answer}]
    # Instruction for generating bounding box
"""
Your task is to generate the bounding boxes of objects mentioned in the caption.
The image is size 512x512. The bounding box should be in the format of (x, y, width, height). Please considering the frame of reference of caption and direction of reference object if possible. If needed, you can make the reasonable guess.
"""
\end{lstlisting}

\end{document}